\documentclass[runningheads]{llncs}

\usepackage{eccv}

\usepackage{multirow}
\usepackage{microtype} 
\usepackage[export]{adjustbox}
\usepackage{eccvabbrv}

\usepackage{graphicx}
\usepackage{booktabs}
\usepackage{color}
\definecolor{light}{rgb}{0.3, 0.3, 0.3}
\def\light#1{{\color{light}#1}}
\def\lightbf#1{\textbf{{\color{light}#1}}}
\newcommand{\bft}[1]{\textbf{#1}}

\newcommand{\nparagraph}[1]{
  \paragraph{\normalfont\bfseries #1}%
}

\let\titleold\title
\renewcommand{\title}[1]{\titleold{#1}\newcommand{\thetitle}{#1}}
\def\maketitlesupplementary
   {
   \newpage
   \begin{center}
      \Large
      \textbf{Supplementary Material} \\
      \vspace{1.0em}
   \end{center}
   }
\usepackage[accsupp]{axessibility}  
\usepackage{makecell} 

\usepackage{hyperref}

\usepackage{orcidlink}

\begin{document}

\title{SPVLoc: Semantic Panoramic Viewport Matching for 6D Camera Localization in Unseen Environments} 

\titlerunning{SPVLoc: Semantic Panoramic Viewport Matching for 6D Localization}

\author{Niklas Gard\inst{1,2}\orcidlink{0000-0002-0227-2857} \and
Anna Hilsmann\inst{1}\orcidlink{0000-0002-2086-0951} \and
Peter Eisert\inst{1,2}\orcidlink{0000-0001-8378-4805}}

\authorrunning{N.~Gard \etal}

\institute{Fraunhofer Heinrich Hertz Institute, HHI, Berlin, Germany \\
\email{\{first.last\}@hhi.fraunhofer.de} \and
Humboldt University of Berlin, Germany }

\maketitle

\begin{abstract}
In this paper, we present SPVLoc, a global indoor localization method that accurately determines the six-dimensional (6D) camera pose of a query image and requires minimal scene-specific prior knowledge and no scene-specific training. Our approach employs a novel matching procedure to localize the perspective camera's viewport, given as an RGB image, within a set of panoramic semantic layout representations of the indoor environment. The panoramas are rendered from an untextured 3D reference model, which only comprises approximate structural information about room shapes, along with door and window annotations. We demonstrate that a straightforward convolutional network structure can successfully achieve image-to-panorama and ultimately image-to-model matching. Through a viewport classification score, we rank reference panoramas and select the best match for the query image. Then, a 6D relative pose is estimated between the chosen panorama and query image. Our experiments demonstrate that this approach not only efficiently bridges the domain gap but also generalizes well to previously unseen scenes that are not part of the training data. Moreover, it achieves superior localization accuracy compared to the state of the art methods and also estimates more degrees of freedom of the camera pose. Our source code is publicly available at: \url{https://fraunhoferhhi.github.io/spvloc}.
\end{abstract}    
\section{Introduction}
\label{sec:intro}
Indoor camera localization is a fundamental challenge of computer vision, aiming to precisely determine a camera's position and orientation within an indoor environment. 
Applications include navigation \cite{Taira2018, Wang2015}, augmented reality (AR) \cite{Acharya2019bim}, building management, and building digitization \cite{Liu2015}.
Current methods often depend on extensive scene-specific data, acquired through manual creation of databases with localized images, depth information, or point clouds \cite{Xia2022, Taira2018, Kendall2015}. However, this limits their applicability, especially in situations where exploring the building is impractical or data acquisition is costly and time-consuming. 

\begin{figure}[!t]
\centering
    \includegraphics[width=.69\linewidth]{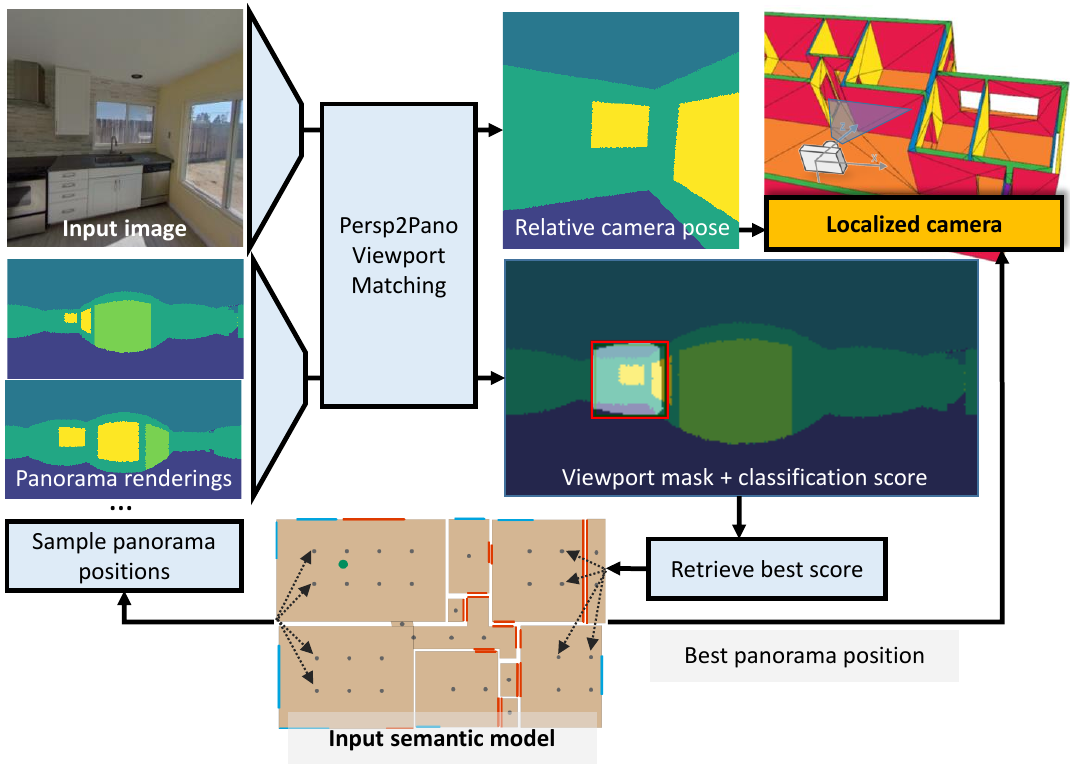}
   \caption{\textbf{SPVLoc for 6D indoor localization.} Our method calculates the indoor 6D camera pose by determining the image position and orientation relative to synthetic panoramas. The best panoramic match is found through semantic viewport matching.}
    \label{fig:overview}
\end{figure}

Recent methods incorporate synthetic data during training, but these localization models require retraining or fine-tuning for new scenes using detailed BIM (Building Information Modeling) models \cite{Acharya2019, Acharya2023}. 
The adoption of the Structured3D format \cite{Sheng2020} streamlines and standardizes 3D structure annotation by simplifying complex models into their most fundamental geometric representations. Methods that match a query image to this simplified structural representation exhibit robustness in unexplored, dynamic environments \cite{Howard2021, Min2022}. 
However, existing methods \cite{Min2022, Chen2024} mainly focus on 2D perspective camera localization with respect to floor plans, missing two degrees of freedom for rotation and one degree of freedom for position. Also, some methods solely rely on panoramic queries \cite{Howard2021, Howard2022, Kim2024}. These queries are typically captured under controlled conditions, involving a tripod-mounted camera with a precise horizontal alignment, and a fixed camera height \cite{Howard2021}. However, in AR applications or scenarios where archived photos must be accurately registered, e.g.~to annotate visible building details into a digital building model, idealized capture conditions cannot always be guaranteed. In such cases, the demand for precise 6D localization becomes imperative.

In this paper, we present SPVLoc, a novel method for indoor 6D camera localization in unseen environments. It represents a 3D scene with a set of synthetically generated omnidirectional panorama images depicting fundamental semantic labels such as door, window, wall, floor, and ceiling. The core idea involves correlating the deep embedding of a perspective RGB camera image with the embeddings of panoramas to estimate the camera viewport – the precise region visible to the camera within the panorama.
Leveraging a RetinaNet-like architecture \cite{Lin2017}, the 2D-bounding box around the viewport is predicted and classified. Retrieval of the closest panorama is performed by selecting the panorama with the globally highest classification score. The exact 6D pose is then found via relative pose estimation with an MLP starting from the panorama's position. 
Our method offers several advantages. It relies solely on a minimalist 3D semantic building model \cite{Sheng2020}, which could even be automatically extracted from simple 2D plans\cite{Cambeiro2023,Lv2021,Park2021}. By bridging the domain gap between semantic representation and real images, it facilitates matching in unseen environments. Our approach (\cref{fig:overview}) learns wide baseline relative pose estimation and accurately predicts poses with few reference renderings.

In our experiments, we evaluate our method using publicly available datasets, including \textit{Structured3D} \cite{Sheng2020} and \textit{Zillow Indoor} \cite{Cruz2021}, encompassing both photorealistic synthetic data and real-world data. 
By sampling perspective query images from equirectangular panoramas, we efficiently augment the dataset, generating an almost infinite amount of training data. Compared to the state of the art, our method excels in estimating all degrees of freedom of the camera pose, enabling precise localization even in uncontrolled recording conditions. To summarise, we make the following contributions.

\begin{enumerate}
\setlength\itemsep{0em}
\item We introduce a model-based 6D camera pose estimation system for unseen indoor environments without the need for scene-specific training.
\item We present a novel perspective-to-panoramic image matching concept with high retrieval accuracy even under wide camera baselines.
\item Demonstrating superior localization accuracy compared to the state of the art, our method estimates more degrees of freedom concurrently.
\end{enumerate}
\section{Related Work}
\nparagraph{Indoor Camera Localization.}
Indoor camera localization is a longstanding challenge in the computer vision community and in recent years, with the advance of neural networks, several deep learning based approaches have been proposed. Typically, these methods rely on prior scene acquisition to built a detailed scene representation and estimate a relation to a query image. Some methods use a 3D model generated via Structure-from-Motion \cite{Agarwal2011} from registered camera images of the indoor environment and calculate the pose from 2D-3D correspondences, which might be either sparse \cite{Liu2017, Xia2022} or dense \cite{Taira2018}. Alternatively, encoding the entire scene within a neural network and performing regression to determine the absolute pose is a frequently used strategy \cite{Naseer2017}. 

By leveraging synthetic renderings from a detailed 3D BIM model, Acharya \etal eliminated the need for real-world reference data. Their approaches achieve real-time indoor localization \cite{Acharya2019} and an accuracy of approximately 1 meter \cite{Acharya2022} but are limited to the building of their origin, lacking generalizability to other structures.
Liu \etal \cite{Liu2015} pioneered floor plan-based image localization, predicting the facing wall within a cuboidal layout without estimating a full 3D pose.
Howard \etal achieved 2D panorama localization in unseen scenes by aligning inferred room layouts with panoramic reference renderings \cite{Howard2021} or encoded floor plans \cite{Howard2022}. 

Several methods estimate the 2D position and 1D rotation of a perspective camera relative to a floor plan. PF-net \cite{Karkus2018} uses a recurrent neural network optimized for a differentiable particle filter, designed for sequential updating but suboptimal for single images \cite{Min2022}. F\textsuperscript{3}Loc \cite{Chen2024} improves sequential 2D localization against a floorplan by integrating multiview geometry cues.
LASER  \cite{Min2022} localizes the camera with respect to a semantic floor plan, and considers both perspective and panoramic queries. The rendering code book scheme efficiently encodes floor plan information but is limited to 2D data, which constrains the network's ability to estimate the remaining degrees of freedom. 

To sum up, existing methods either depend on accurate scene specific data for a detailed 3D representation or do not estimate the 6D camera pose. In contrast, our method neither depends on real-world reference images, a detailed 3D model, or scene specific training but is still able to estimate the 6D perspective camera pose by introducing a new cross-domain panoramic-to-perspective image registration approach between RGB images and semantic panoramas. 

\nparagraph{Relative Pose Regression.}
Relative pose regression (RPR) aims at estimating the relative motion between a query image and reference images with known poses. These frameworks, relying on image comparisons, often enhance generalizability. Typically, they directly utilize localized examples at inference time \cite{Ding2019}.

Retrieval-based methods like NetVLAD \cite{Arandjelovic2016} and DenseVLAD \cite{Torii2015} start by identifying  the most similar images in a database, providing a starting point for RPR. PixLoc \cite{Sarlin2021} predicts multiscale deep features for direct alignment of reference and query image with a CNN and uses geometric optimization for calculating the pose. 
Turkoglu \etal \cite{Turkoglu2021} employ a GNN to re-localize cameras with respect to multiple frames. A graph links the query image to training counterparts, and a GNN refines representations for a consistent camera pose estimate. Our method shares similarities with \cite{Turkoglu2021} but goes further by addressing the domain gap between simplified synthetic and real data, and by  incorporating multimodal matching between panoramic reference and perspective query images.

\nparagraph{Perspective-to-Panorama Matching.}
Matching between perspective query and panoramic reference images is a crucial aspect of localization, yet it has received limited attention in the literature. Most systems avoid multimodal matching and instead either match perspective crops to perspective querys \cite{Arandjelovic2016, Torii2015} or panoramic to panoramic images \cite{Zhang2021}. 

While the past discussed geometric fundamentals for integrating catadioptric and perspective camera systems into hybrid stereo setups, including concepts like epipolar constraints \cite{Sturm2002} and calibration \cite{He2012}, these ideas have not yet found their way into contemporary AI-based localization systems. In the domain of CNNs and panoramic image processing, distortion-aware convolutional filters were proposed by Tateno \etal ~\cite{Tateno2018} and further refined by Fernandez \etal \cite{Fernandez2020}, particularly for indoor equirectangular images. Unlike traditional convolutions, these distortion-aware filters target the differences betweeen perspective and panoramic images, making them particularly valuable for cross-modal matching.
 
Recently, perspective-to-panoramic matching advanced, due to sliding window approaches in feature space \cite{Orhan2021, Shi2023}. It  improved visual recognition compared to methods that simply sampled perspective reference images from panoramas \cite{Zamir2010}. Orhan~\etal \cite{Orhan2022} explored leveraging semantic segmentation similarity to address appearance variations between reference and query images.
Our approach enhances perspective-to-panorama matching using virtual scene renderings as panoramic references. Inspired by multimodal zero-shot retrieval \cite{Janik2021} and object detection, particularly Mercier \etal.'s work \cite{Mercier2021}, it utilizes the panorama as a global context for query matching.
\section{Method}
SPVLoc conducts 6D indoor localization of a 2D RGB query image within a simple semantic textureless 3D scene model. The model adheres to the Structured3D annotation format \cite{Sheng2020}, containing walls, doors, windows, ceiling, and floor as planes. It can be derived from straightforward from building models  \cite{Sheng2020}, or from floor plans using contemporary AI tools \cite{Cambeiro2023,Lv2021,Park2021}.
SPVLoc first estimates the image's viewport within rendered panoramas from the semantic scene model using cross-domain image-to-panorama matching (\cref{sec::viewport}). It then ranks these predictions based on classification accuracy (\cref{sec::inference}) and performs relative 6D pose regression with respect to the best-matching reference panorama (\cref{sec::relative_pose}).

\begin{figure*}[!t]
    \includegraphics[width=\linewidth]{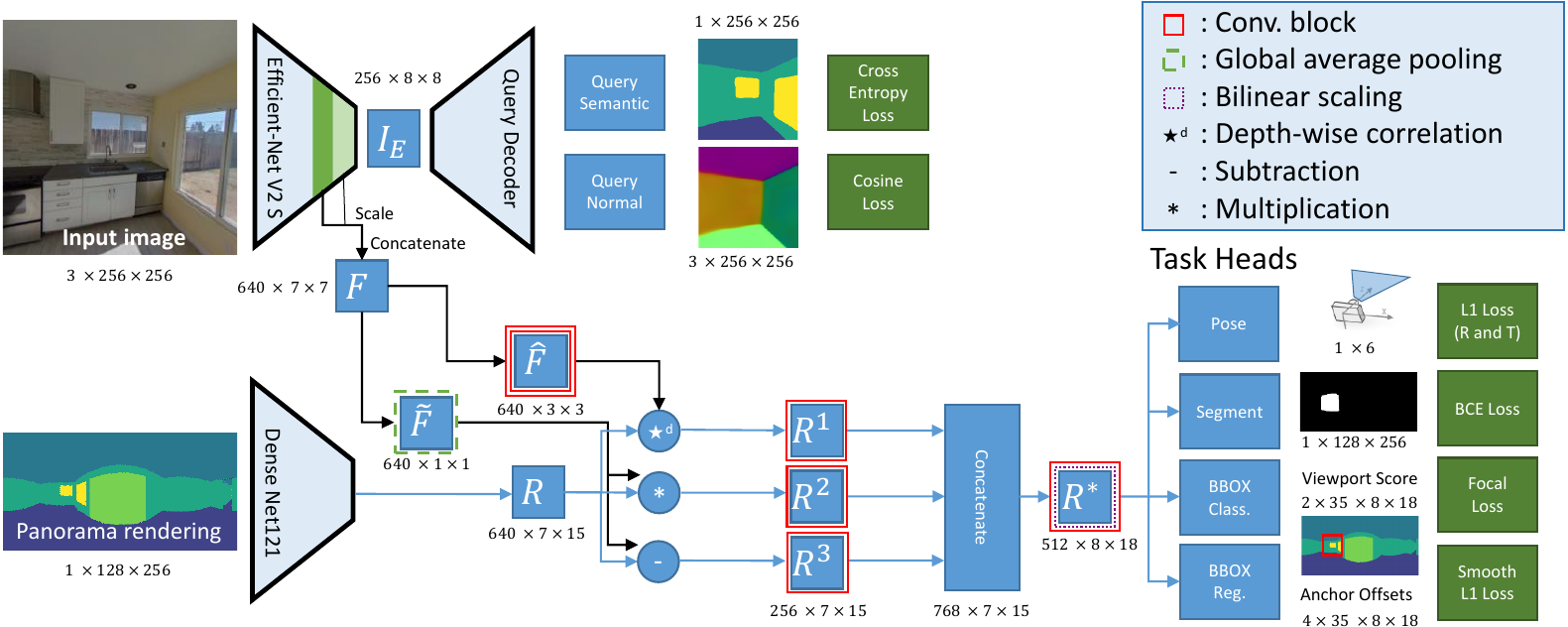}
   \caption{\textbf{Network architecture overview.} The information from the query branch is fed into the panorama branch by depth-wise correlation. Three task heads predict the corresponding viewport, one task head predicts the relative pose offset.}
    \label{fig:view_correlation}
\end{figure*}

\subsection{Semantic Panoramic Viewport Matching}
\label{sec::viewport}

\nparagraph{Problem Definition.} 
Our objective is to precisely estimate the 6D camera pose denoted as $P$ for a given query RGB image $I_q$ relative to a semantic 3D model $S$. To achieve this, we reformulate the 6D indoor localization of $I_q$ within $S$ as a cross-domain image-to-panorama matching problem. The model is represented by a set of $N$ equirectangular reference layouts $L = \{L_i~|~i = 1, \ldots N\}$ consisting of semantic labels in a single channel. Each rendering comprehensively captures the entire environment around a specific vantage point. The viewport $V$ is a distinct \textit{cutout} section from a panorama, defining the exact region a camera positioned nearby can see. The perspective distortion of $V_{P_i}$ contains information about the relative pose offset $P_i$ between $L_i$ and $I_q$. Scoring and ranking viewports helps in choosing the optimal panorama $L^\star$.
The absolute pose $P$ is then formed by concatenating $P^\star$ with $P_i$. Identifying the viewport and the corresponding pose offset poses several challenges. First, with increasing distance between the two cameras the perspective distortion between image and panorama increases. Second, the matching involves a severe domain gap as a photographic RGB image is matched against a rendered semantic panorama.

\nparagraph{Viewport Prediction.} 
Semantic panorama reference renderings are created utilizing a ray-tracing-based renderer \cite{Li2018}. The viewport mask $V_P$ is a binary image that signifies the visibility of points within the equirectangular image from the view of the perspective camera. The calculation involves projecting all 3D points from the panorama in the perspective view. The projection is performed based on the known underlying layout during the training phase. The mask $V_P$ is determined by identifying pixels projecting into the perspective image, taking into account the 3D camera pose, and excluding those covered by occlusion.

Additionally, the 2D bounding box $v_{bb}$ is the minimal axis-aligned frame around $V_P$. This bounding box is circularly defined within the 360$^{\circ}$ panorama, allowing the overlap on the right image side to re-enter on the left side of the panorama. Each predicted bounding box is associated with a confidence score $c_{bb}$, which quantifies the probability of the bounding box being a valid match. 
The network's objective is to predict $V_P$, $v_{bb}$ and $c_{bb}$ for each pair of input semantic panorama and query RGB image. The optimal viewpoint corresponds to the predicted bounding box with the highest overall confidence score, $c_{bb}$, when presented with a set of panorama images.

\nparagraph{Network Architecture.} 
Our matching is inspired by the object detection strategy of DTOID \cite{Mercier2021}. \Cref{fig:view_correlation} illustrates the network architecture. Initially, $I_q$ is encoded with an \textit{EfficientNet-S} backbone. The features obtained after the fourth and the last downsampling layer are preserved. Subsequently, both outputs are rescaled to a tensor of spatial size of $7 \times 7$ and concatenated to a tensor $F$. 

$L$ is encoded using a \textit{DenseNet-121} backbone, extending up to the fifth downsampling block to extract features $R$. The information extracted from the input image guides the panorama branch to estimate the viewport mask and bounding box. This guidance is achieved through the application of a depth-wise correlation technique (denoted as $\mathbf{\star_d}$) \cite{Mercier2021, Ammirato2018}. 
$F$ is processed through two consecutive convolution blocks, each incorporating convolution, batch normalization, and ELU activation, resulting in a set of $3 \times 3$ filters $\hat{F}$ used during the depth-wise correlation $R\mathbf{\star_d}\hat{F}$. 

Furthermore, $F$ undergoes compression via global max pooling and is both multiplied and subtracted from $R$, resulting in three feature tensors with injected image information. The features are individually processed with an additional convolution block and then concatenated. The concatenated features further undergo a final convolution block, leading to the correlated features $R^\star$.

A decoder estimates the viewport of the perspective camera projected in the layout of the omnidirectional panorama. The viewport in the panorama is estimated as a bounding box with a \textit{BBox Class.}~and \textit{BBox Reg.}~head, similar to RetinaNet \cite{Lin2017}. 
The \textit{BBox Class.}~head's output is optimized using focal loss $\mathcal{L}^{VP}_1$, while the \textit{BBox Reg.}~head is optimized with a \textit{smooth} $l_1$ \textit{loss} $\mathcal{L}^{VP}_2$. 
The \textit{BBox Class.}~provides a score for a given estimation, ensuring that the bounding box with the highest probability corresponds to the best estimate. 
Additionally, $V_P$ representing the exact viewport of the perspective camera is estimated with the aid of another convolutional decoder head and optimized with binary cross entropy loss $\mathcal{L}^{VP}_3$. 

\nparagraph{Perspective Supervision.} 
The encoded image features $I_E$ are connected to a convolutional decoder, outputting the underlying semantic and normal map to ensure that features closely corresponding to the image content are extracted. The semantics are learned with cross entropy loss $\mathcal{L}^{VP}_4$ and the normals are learned with cosine loss $\mathcal{L}^{VP}_5$.

\subsection{Feature-Correlation-based Pose Regression}
\label{sec::relative_pose}
The correlated features, denoted as $R^\star$, encode information about the viewport of $I_q$ within a $L_i$, but should also lead to relative pose offset $P_i$ between them. Given that $L$ is synthetically generated, the precise position of $L_i$ is known. Therefore, determining the relative position of $I_q$ in relation to $L_i$ is equivalent to establishing its absolute position. All rendered panoramas exhibit identity orientation, due to horizontal alignment and a northward orientation. Consequently, the rotation delta directly corresponds to the absolute rotation of the camera.

\begin{figure}
\centering
    \includegraphics[width=.65\linewidth]{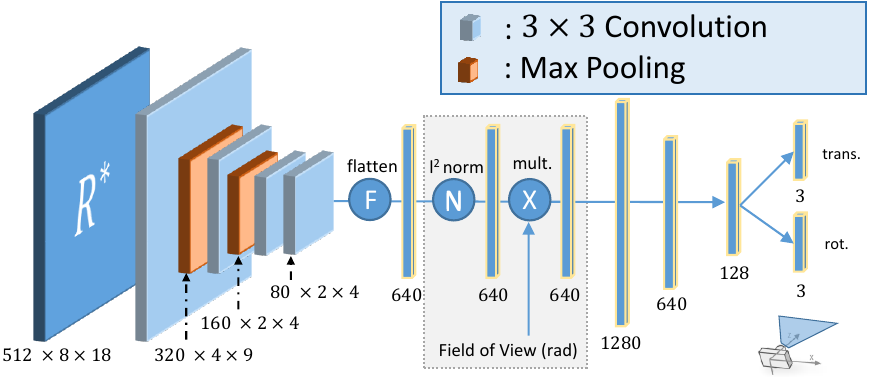}
   \caption{The \textit{Pose} head utilizes convolutions, an MLP, and a FoV-based side input.}
    \label{fig:pose_head}
\end{figure}

An additional \textit{Pose} head processes $R^\star$ to estimate the relative pose offset of the camera concerning a panorama rendering (\cref{fig:pose_head}). We first utilize three convolution blocks, whereby the first two are followed by a max-pooling layer, to compress the features.
The flattened features form an 1D embedding $r^\star \in \mathbb{R}^C$. Here, $C = 640$ for a panorama of dimensions $256 \times 128$. 
Optionally, to handle different camera field of views during training and testing, $r^\star$ is normalized and multiplied by the horizontal field of view angle in radians. The result is fed into a four-layer MLP, producing the 3D offset to the query image location $\hat{t}{i} \in \mathbb{R}^{3}$ and a rotation given as the logarithm of the unit quaternion $\hat{r}{i} = \log(\hat{q})_{ij} \in \mathbb{R}^3$ \cite{Turkoglu2021}.

The MLP is evaluated using two loss functions. $\mathcal{L}^{P}_1$ assesses the precision of the relative translation, and $\mathcal{L}^{P}_2$ evaluates the accuracy of the final rotation estimation. Both losses are computed using the $l_1$ distance.

\subsection{Optimization}
The training loss is the sum of the two losses of the MLP and the five losses of the viewport estimation. With seven losses, our multi-task learning has to be balanced by using a different weight for each loss. As in \cite{Kendall2018, Kendall2017}, we use an exponential mapping and weight multiple loss functions by considering the homoscedastic uncertainty of each task and learned weighting factors. The final loss is given as:
\begin{equation}
\mathcal{L} = \sum_{i=1}^{2} \mathcal{L}^{P}_i e^{-\beta_i} + \beta_i + \sum_{i=1}^{5} \mathcal{L}^{VP}_i e^{-\gamma_i} + \gamma_i
\end{equation}

\subsection{Inference}
\label{sec::inference}
During inference, panorama positions are determined from a fixed 2D grid superimposed on the floor plan, with virtual cameras set at a constant height denoted as $h_{pano}$, and the grid's dimensions specified as $xy_{pano}$. Virtual cameras are exclusively sampled above floors to ensure valid reference images inside rooms. Setting $xy_{pano}$ as large as possible is advantageous in order to minimize the number of references, given that feature correlation and bounding box regression are executed for every panorama. 

For pose estimation, we select reference positions with the top-$n$ highest bounding box classification scores. The absolute pose is determined by the result of the \textit{Pose} head, where the selected reference position is added to the translation. To eliminate the necessity for rendering during inference, we propose pre-generating reference panoramas for a floor plan. Additionally, the encoding of these panoramas can be precomputed. This ensures that only the encoded features $R$ need to be stored, significantly streamlining the inference process.

\nparagraph{Refinement.} 
Without altering the network architecture or training process, SPVLoc can enhance an estimated pose by rendering a new reference panorama at the estimated position and re-executing the relative pose regression. The smaller step to the actual target position in this second iteration simplifies the problem, allowing for increased precision in the refined pose estimation.
\section{Experiments}
\label{sec:experiments}
To validate our method we conduct several experiments and make use of two publically available and frequently used datasets. 

\nparagraph{Structured3D (S3D)~\cite{Sheng2020}} consists of 3500 near-photorealistic professionally modeled indoor environments, each annotated with ground truth 3D structure information. It encompasses 21835 panoramic images, each provided in both furnished and unfurnished configurations. 
The authors define a unified representation that describes scenes using geometric primitives such as planes, lines, junction points, and semantic annotations including floor, ceiling, wall, door, and window classes.

\begin{figure}[t]
\centering
    \includegraphics[width=\linewidth]{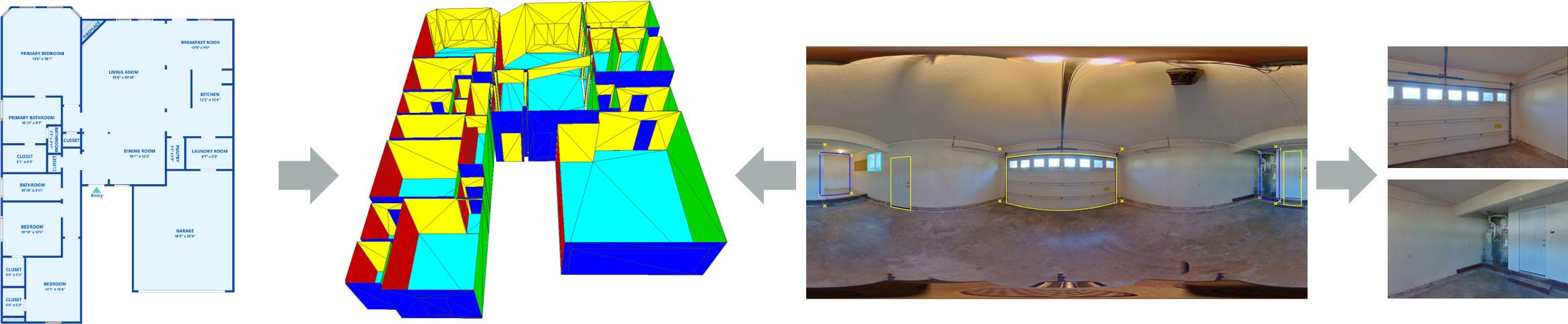}
   \caption{\textbf{ZinD data preparation.} Annotations generate 3D reference models (left), while resampled panoramas create perspective train and test images (right).}
    \label{fig:zillow}
\end{figure}

\nparagraph{Zillow Indoor (ZInD) \cite{Cruz2021}} includes 67,448 panorama images from 1,575 unfurnished residential homes, globally aligned and registered to floor plans. The 3D annotations of panoramas within a scene are adequate for transforming them into the S3D representation format. Before training, we transform all data to that format. \Cref{fig:zillow} shows an example reference model generated from ZInD annotations.
Unlike S3D, ZInD includes room openings such as corridors connected to rooms without doors, enhancing structural variation in semantic renderings.

\smallskip
For both datasets, we employ the official train-test scene split. To enhance control over the field of view and angle variation, we sample perspective test images from the RGB panoramas. They are registered in the 3D model, so the semantics are known for each perspective image. We compile test-sets suitable for 6D pose estimation and structured evaluation.
The test sets consist of square perspective images with a viewing angle of $60$, $90$, and $120^\circ$, respectively, along with variations in roll and pitch, and random yaw angles for ZinD and S3D. To ensure meaningful views, we apply two criteria on the sampled yaw. First, the images contain a minimum of three semantic classes, like wall-floor-ceiling or wall-window-floor. Second, the camera faces inside the room rather than towards the next wall. The criteria are applied to the $60^\circ$ images (4420 for ZiND and 1133 for S3D). Subsequent variations in opening angles and roll-pitch configurations are generated from this starting point. 
\subsection{Details}
For evaluation, we report the fraction of frames localized correctly within a threshold $\tau$. We also list the median rotation and translation error, as well as the fraction of frames localized within  $\tau$ and an additional angle threshold $\upsilon$. In the retrieval process, the top-3 best matches are selected, and we report recall metrics for both top-1 and the best among the top-3. 

\nparagraph{Training.} For the state of the art comparison, we train a model with a variable field of view, while other experiments use a model trained with a fixed opening angle of $90^\circ$. The model with variable viewing angles randomly samples views between $45$ and $135^\circ$ during training. 
We consistently render $s$ randomly positioned panoramas around the query within a radius of $\pm r_1$ in $xy$ direction and $\pm r_2$ in $z$ (upward) direction. For a single query, the viewport and relative pose offset is therefore calculated against multiple references. In our experimental setup, we set $s$ to 4, $r_1$ to 1.4 meters, and $r_2$ to 30 centimeters. Additionally, one random \textit{negative} example is generated outside $r_1$, likely in a different room, with the target to enhance the network's ability to distinguish subtle room differences. Images are sampled with random yaw and $\pm 10^\circ$ random pitch and roll angle. The batch size is set to 40, comprising 40 query images and 200 panoramas, and the training is performed on a single NVIDIA A100 GPU. Query images with fewer than three semantic classes are disregarded during the loss calculation process. Training runs for approximately 42,000 steps with an initial learning rate of $2.5 \times 10^{-4}$, halved two times during training.

\nparagraph{Testing.} During testing, panoramas are sampled on a $1.2 \times 1.2$ m grid, requiring significantly fewer samples than other rendering-based approaches such as LaLaLoc \cite{Howard2021}, which uses a $0.5 \times 0.5$ m grid. Additionally, for a scene, all panoramic views can be encoded initially, and during inference, only the query image needs to be encoded. If not otherwise noted the studies are conducted on ZInD with fixed 90 degree FoV, reporting the 3D rotation and translation error.

\subsection{Comparison with the State of the Art}
\begin{table*}[t!]
    \caption{\textbf{Comparison with state-of-the-art baseline on ZInD and S3D}: Median translation (terr\textsubscript{xy}) in cm, rotation error (rerr\textsubscript{yaw}) in deg. for instances localized under 1m. Reporting recall at various translation accuracy levels, recall for inliers ($<$1m and $<$30°), and top-3 recall at 1m. \textit{Ours\textsuperscript{\textdagger}} refers to our model tested with a slightly different testset, incorporating pitch and roll rotation variation ($\pm 10^\circ$).} 
    \renewcommand{\arraystretch}{1.1}
    \centering
    \resizebox{\linewidth}{!}{%
        \begin{tabular}{ccccccccc|ccccccc} 
\toprule
 &  & \multicolumn{7}{c|}{ZInD} & \multicolumn{7}{c}{Structured3D \, (Furnishing-Level : Full)} \\ 
\hline
Query & Method & \begin{tabular}[c]{@{}c@{}}\textless{}1m med\vspace{-0.3em}\\terr (cm)\end{tabular} & \begin{tabular}[c]{@{}c@{}}\textless{}1m med\vspace{-0.3em}\\rerr (deg)\end{tabular} & \begin{tabular}[c]{@{}c@{}}10cm\vspace{-0.3em}\\(\%)\end{tabular} & \begin{tabular}[c]{@{}c@{}}50cm\vspace{-0.3em}\\(\%)\end{tabular} & \begin{tabular}[c]{@{}c@{}}1m\vspace{-0.3em}\\(\%)\end{tabular} & \begin{tabular}[c]{@{}c@{}}1m \& 30°\vspace{-0.3em}\\ (\%)\end{tabular} & \begin{tabular}[c]{@{}c@{}}top-3\vspace{-0.3em}\\1m (\%)\end{tabular} & \begin{tabular}[c]{@{}c@{}}\textless{}1m med\vspace{-0.3em}\\terr (cm)\end{tabular} & \begin{tabular}[c]{@{}c@{}}\textless{}1m med\vspace{-0.3em}\\rerr (deg)\end{tabular} & \begin{tabular}[c]{@{}c@{}}10cm\vspace{-0.3em}\\(\%)\end{tabular} & \begin{tabular}[c]{@{}c@{}}50cm\vspace{-0.3em}\\(\%)\end{tabular} & \begin{tabular}[c]{@{}c@{}}1m\vspace{-0.3em}\\(\%)\end{tabular} & \begin{tabular}[c]{@{}c@{}}1m \& 30°\vspace{-0.3em}\\(\%)\end{tabular} & \begin{tabular}[c]{@{}c@{}}top-3\vspace{-0.3em}\\1m(\%)\end{tabular} \\ 
\hline
\multirow{3}{*}{\begin{tabular}[c]{@{}c@{}}Perspective\\60° FoV\end{tabular}} 
& LASER  \cite{Min2022}     & 33.53 & \bft{1.03} & 4.05 & 47.24 & 64.52 & 63.53 & 78.64   & 17.60 & \bft{0.84} & 16.46 & 63.22 & 68.39 & 67.34 & 86.08 \\
& Ours & \bft{17.16} & 1.50 & \bft{21.04} & \bft{81.33} & \bft{86.76} & \bft{86.09} & \bft{94.12} & \bft{12.25} & 1.21 & \bft{34.94} & \bft{87.13} & \bft{89.49} & \bft{88.79} & \bft{96.15} \\
& \light{Ours\textsuperscript{\textdagger}}  & \light{17.77} & \light{1.63} & \light{19.59} & \light{81.13} & \light{87.10} & \light{86.43} & \light{94.86} & 
\light{13.31} & \light{1.24} & \light{30.82} & \light{84.68} & \light{88.18} & \light{87.48} & \light{95.80} \\
\hline

\multirow{3}{*}{\begin{tabular}[c]{@{}c@{}}Perspective\\90° FoV\end{tabular}} 
& LASER \cite{Min2022}      & 26.28 & \bft{0.77} & 8.69 & 67.01 & 80.90 & 80.25 & 89.73    & 14.60 & \bft{0.67} & 23.47 & 72.77 & 76.80 & 76.18 & 91.94 \\
& Ours & \bft{13.63} & 1.34 & \bft{30.57} & \bft{85.59} & \bft{90.84} & \bft{90.34} & \bft{96.36} & \bft{9.17} & 0.83 & \bft{50.09} & \bft{89.67} & \bft{92.12} & \bft{91.86} & \bft{97.37} \\
& \light{Ours\textsuperscript{\textdagger}}  & \light{14.25} & \light{1.37} & \light{29.59} & \light{85.90} & \light{91.33} & \light{90.95} & \light{96.97}  & 
\light{10.09} & \light{0.91} & \light{45.62} & \light{89.32} & \light{92.12} & \light{91.94} & \light{96.85} \\
 
\hline
\multirow{3}{*}{\begin{tabular}[c]{@{}c@{}}Perspective\\120° FoV\end{tabular}} 
& LASER \cite{Min2022}      & 23.10 & \bft{0.78} & 12.69 & 75.09 & 85.63 & 85.20 & 92.38  & 12.90 & \bft{0.76} & 29.07 & 77.76 & 82.14 & 81.35 & 94.13 \\
& Ours & 
\bft{14.78} & 1.45 & \bft{29.05} & \bft{85.14} & \bft{91.72} & \bft{91.27} & \bft{96.92} & \bft{11.69} & 0.87 & \bft{39.49} & \bft{88.00} & \bft{92.12} & \bft{91.94} & \bft{97.37} \\

& \light{Ours\textsuperscript{\textdagger}}  & 
\light{15.13} & \light{1.53} & \light{28.12} & \light{86.13} & \light{92.71} & \light{92.35} & \light{97.24} & \light{11.77} & \light{0.88} & \light{38.18} & \light{89.67} & \light{93.08} & \light{92.91} & \light{97.55} \\
\bottomrule
\end{tabular}
    }

    \label{tab:sota}
\end{table*}
\label{sec:experiments_stoa_loc}
To our knowledge, SPVLoc is the first method to perform 6D localization in unseen scenes with respect to semantic 3D models. The state of the-art method for scene-independent global localization, LASER \cite{Min2022}, takes perspective views as input and matches them against a semantic floor plan. Hence, it estimates only two position and one rotation degree of freedom. Nonetheless, it serves as a strong baseline to assess the localization accuracy of our method.

\Cref{tab:sota} shows the results on ZInD and S3D. Given that we compare against a 2D method, our evaluation also includes recall based on 2D distance and angular error derived from the yaw angle. The yaw angle is extracted from the estimated rotation matrix. Notably, SPVLoc without refinement outperforms LASER in terms of localization accuracy, exhibiting higher recall across all categories, an increased inlier rate, and significantly reduced median translation error. In terms of 2D rotation, our method is slightly less precise. This is because LASER estimates yaw directly while we estimate a full rotation matrix. Estimating camera height and roll and pitch angle is an advantage of our method. In each last row (\textit{Ours\textsuperscript{\textdagger}}), we evaluate our method on an alternate test set with roll and pitch variations ($\pm 10^\circ$). The results suggest that these variations do not substantially impact the accuracy of our method for 2D localization. 

Regarding further methods, LASER has showcased a significant performance improvement over PF-net \cite{Karkus2018} and non-AI-based Monte-Carlo Localization (MCL)\cite{Dellaert1999}. PF-net, originally designed for sequential time updating, does not perform well in single-query localization scenarios. MCL neglects semantic annotations, uses LIDAR instead of images, which also has been shown to drastically reduce performance \cite{Min2022}. Our substantial improvement over LASER therefore also illustrates a performance gain over aforementioned methods. Another experiment using the pre-rendered perspective images of S3D, including a comparison to F\textsuperscript{3}Loc \cite{Chen2024}, is part of the supplementary material.

\subsection{Performance Studies}\label{sec:experiments_performance}

\nparagraph{Ablation Study.}
\begin{table}[t!]
\caption{\textbf{Ablation study.} SPVLoc variations compared on ZInD with $90^\circ$ FoV. In contrast to \cref{tab:sota}, terr denotes 3D position errors, and rerr denotes 3D rotation errors.}
\renewcommand{\arraystretch}{1.1}
\centering
\resizebox{.7\columnwidth}{!}{%
    \begin{tabular}{cclcccccccc}
\toprule
 Abl. Type & ~ & Method & \begin{tabular}[c]{@{}c@{}}\textless{}1m med\vspace{-0.3em}\\terr (cm)\end{tabular} & \begin{tabular}[c]{@{}c@{}}\textless{}1m med\vspace{-0.3em}\\rerr (deg)\end{tabular} & \begin{tabular}[c]{@{}c@{}}10cm\vspace{-0.3em}\\(\%)\end{tabular} & \begin{tabular}[c]{@{}c@{}}50cm\vspace{-0.3em}\\(\%)\end{tabular} & \begin{tabular}[c]{@{}c@{}}1m\vspace{-0.3em}\\(\%)\end{tabular} & \begin{tabular}[c]{@{}c@{}}1m \& 30°\vspace{-0.3em}\\ (\%)\end{tabular} & \begin{tabular}[c]{@{}c@{}}top-3\vspace{-0.3em}\\1m (\%)\end{tabular} \\
\hline
\multirow{2}{*}{(Ours)} & ~ &base & \textbf{14.31} & \textbf{2.05} & \textbf{26.88} & 86.92 & \textbf{91.90} & \textbf{91.61} & 96.56 \\

 & \light{~+} &\light{refine} & \light{12.32} & \light{1.96} & \light{33.69} & \light{90.84} & \light{92.76} & \light{91.74} & - \\
\hline
\multirow{3}{*}{Loss} & - & query dec. & 15.70 & 2.23 & 21.97 &  85.50 & 90.79 & 90.43 &  96.45\\
& - & view dec. & 15.19 & 2.21 & 23.60 & 85.36 & 90.45 & 89.82 & 96.49 \\
& - & view/query dec. & 16.22 & 2.19 & 21.56 & 84.50 & 89.59 & 89.12 & 96.36\\
\hline
Training & - & neg. samples & 15.06 & 2.28 & 22.60 & 76.65 & 82.19 & 81.74 & 92.99 \\
\hline
\multirow{2}{*}{Arch.} & $\Delta$ & Resnet18 (query) & 15.56 & 2.56 & 20.88 & 86.56 & 90.72 & 90.32 & 96.56 \\
& $\Delta$ & Equiconv (pano) & 14.98 & 2.13 & 25.07 & \textbf{87.44} & 91.70 & 91.31 & \textbf{96.97} \\
\hline
\hline
\multirow{2}{*}{\light{Input}} & \light{~+} & \light{normal} & \light{14.68} & \light{2.28} & \light{25.09} & \light{85.93} & \light{90.84}  & \light{90.23} & \light{96.47}  \\
&  \light{~+} & \light{depth} & \lightbf{14.06} & \light{2.10} & \lightbf{27.60} & \lightbf{88.05} & \lightbf{92.15} & \lightbf{91.79} & \light{96.74} \\
\bottomrule
\end{tabular}

}
\label{tab:spvloc-ablation}
\end{table}
We commence our experimentation with the full network (\textbf{base}) configuration. Subsequently, we modify the network by removing specific components (\cref{tab:spvloc-ablation}). Initially, we test the influence of perspective supervision (\textbf{-query dec.}) and the view segment task head (\textbf{-view dec.}) by excluding them during training. Their exclusion reduces network performance, likely because the network's ability to understand the image content and establish correspondences is diminished. Removing both (\textbf{-view/query dec.}) further reduces accuracy.

We then exclude the negative samples from different rooms during training (\textbf{-neg. samples}). This significantly reduces the amount of correctly localized images. It supports the hypothesis that teaching the network to recognize spaces where there is no match helps to improve the accuracy of predictions. 

To modify the architecture, the image encoder \textit{EfficientNet-S} is replaced with a smaller \textit{ResNet-18}\cite{He2016} ($\mathbf{\Delta}$ \textbf{Resnet18}), resulting in a performance drop. This suggests an increased demand for network capacity to analyze the images. In the panorama encoder, replacing all convolutional layers with \textit{Equiconv} \cite{Fernandez2020} ($\mathbf{\Delta}$ \textbf{Equiconv}), a specialized convolution for panorama images that remaps query locations from Euclidean to spherical space, does not result in a performance gain. This suggests that the capacity of the small \textit{DenseNet-121} is sufficient for analyzing the low-resolution semantic inputs.

Lastly, we incorporate additional input modalities for the panoramas into the panorama encoder. While the addition of world coordinate normals (\textbf{+normals}) did not improve the result, including an extra depth map slightly enhanced the outcome (\textbf{+depth}). This suggests that there are cases where additional information is encoded in the depth that cannot be inferred from semantics alone.

\nparagraph{Qualitative Results.}
\begin{figure*}[!t]
\centering
    \includegraphics[width=\linewidth]{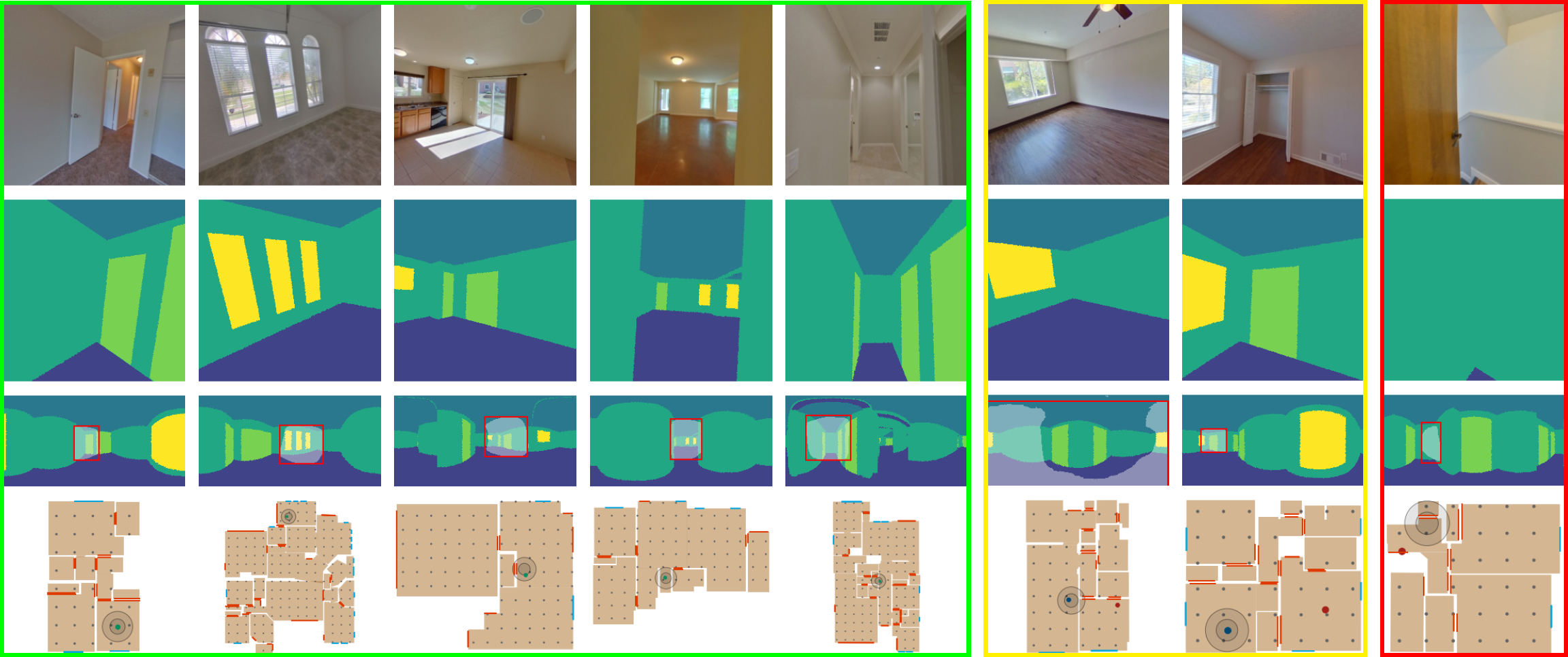}
   \caption{\textbf{Qualitative localization results.} Top to bottom - query, rendering with top-1 estimated pose, panorama with estimated viewport, map. Green box: success for top-1 match. Yellow box: success for top-2 match. Red box: failure case.}
    \label{fig:qualitative}
\end{figure*}
\Cref{fig:qualitative} shows selected SPVLoc outcomes. The top row displays camera images, and the second row shows renderings from the semantic 3D model with estimated top-1 poses. All green boxed images are correctly localized for the top-1 match. In the yellow box, the top-1 match fails, but the top-2 match is accurate. The red box example cannot be localized. Across all examples, a pose with high visual similarity is found. Ambiguous examples arise from similar layouts occurring in multiple rooms. Failure cases are attributed to either images lacking sufficient structural information for matching or outliers in the data, such as those taken on balconies or in dark cellars. 
The third row displays the chosen reference panorama, the selected viewport bounding box corresponding to the camera image, and a semi-transparent overlay representing the estimated viewport mask. Additional detailed visualizations are part of the supplementary material.

\nparagraph{Influence of Sampling Grid Sizes and Refinement.}
Reference positions result from overlaying a grid onto the floorplan, and the grid density significantly affects SPVLoc's efficiency and inference time. Minimizing references reduces rendering effort and requires the model to estimate larger offsets. E.g.~to increase the grid from $0.7m$ to $2m$ reduces the number of samples per $m^2$ by a factor of 10. 
We vary the grid size to analyze its impact on performance. We also experiment with a local grid per room substituting the global grid (\textbf{Adv.}) reducing the risk of missing rooms entirely. The local grid is centered on the room outline and additionally places one camera in the middle of each opening. \Cref{fig:grid_size} shows different grids for an example plan. 
\Cref{tab:grid_size_quantitative} illustrates the relationships among grid size, global and local sampling, and refinement. Refinement notably improves 10cm recall across configurations. For the largest local grid with refinement, the 1m recall only decreases by 1.18\% compared to the baseline configuration. In summary, a trade-off between refinement and dense/sparse scene sampling exists, yet our method consistently delivers strong results with very sparse sampling, making it suitable for large scenes.

\begin{table}[!t]
\begin{minipage}{0.53\linewidth}
\caption{\textbf{Grid size vs. accuracy}: Deviation of 10cm and 1m recall from (boxed) baseline.}
\renewcommand{\arraystretch}{1.1}
\centering
\resizebox{.95\columnwidth}{!}{%
    \begin{tabular}{ccccccccc}
\toprule
 \begin{tabular}[c]{@{}c@{}}Sample\\dist.\end{tabular} 
 & \begin{tabular}[c]{@{}c@{}}Samples\\per $m^2$\end{tabular} 
& \begin{tabular}[c]{@{}c@{}}Adv.\\~\end{tabular} 
 & ~
 & \begin{tabular}[c]{@{}c@{}}~10cm\\(\%)\end{tabular} 
 & \begin{tabular}[c]{@{}c@{}}~10cm\\+~refine\end{tabular} 
 & ~ 
 & \begin{tabular}[c]{@{}c@{}}~1m\\(\%)\end{tabular} 
 & \begin{tabular}[c]{@{}c@{}}~1m\\+~refine\end{tabular} \\
\cline{1-3}\cline{5-6}\cline{8-9}
0.7 &  2.07 & ~ & ~ & $+1.00$ & $+7.49$ & ~ & $+1.86$ & $+2.38$ \\
0.9 & 1.25 & ~ & ~ & $+0.95$ & $+6.99$ & ~ & $+1.79$ & $+2.08$ \\
\cline{5-5}\cline{8-8}
\bft{1.2} & \bft{0.70} & ~ & ~ &\multicolumn{1}{|c|}{$\mathbf{26.88}$} & $+6.81$ & ~ & \multicolumn{1}{|c|}{{$\mathbf{91.90}$}} & $+0.86$ \\
\cline{5-5}\cline{8-8}
1.2 & 0.48 & \checkmark & ~ & $+0.57$ & $+6.95$ & ~ & $-0.63$ & $+0.61$ \\
1.5 & 0.45 & ~           & ~ & $-1.00$ & $+6.63$ & ~ & $-2.47$ & $-1.02$ \\
1.5 & 0.31 & \checkmark & ~ & $-1.04$ & $+6.47$ & ~ & $-2.67$ & $-0.38$ \\
2.0 & 0.26 & ~          & ~ & $-4.19$ & $+3.60$ & ~ & $-9.05$ & $-5.79$ \\
2.0 & 0.20 & \checkmark & ~ & $-1.76$ & $+5.81$ & ~ & $-3.91$ & $-1.18$ \\
\bottomrule
\end{tabular}

}
\label{tab:grid_size_quantitative}
\end{minipage}%
\hfill
\begin{minipage}{0.44\linewidth}
\renewcommand{\arraystretch}{1.1}{
    \includegraphics[width=.95\linewidth]{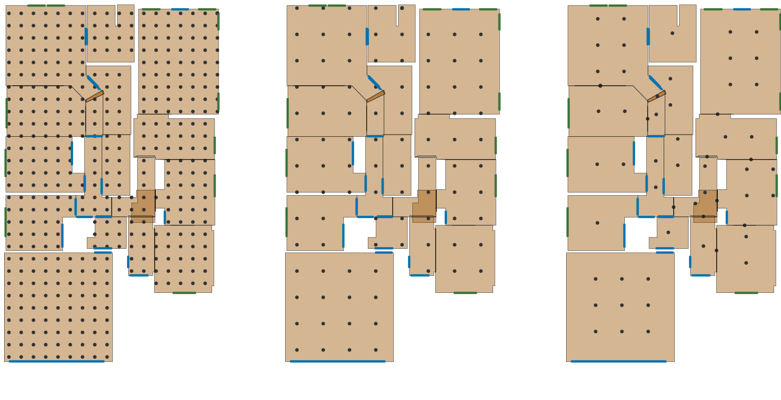}
    \captionof{figure}{Reference positions for global grid of size 0.7m, 1.5m and local grid of size 1.5m (from left to right).}
    \label{fig:grid_size}
}
\end{minipage}
\end{table}

\nparagraph{Influence of Focal Length.}
\Cref{tab:spvloc-fov} depicts the performance difference between a variant trained with a known camera focal length and a focal length-independent network trained with varying focal lengths (ranging from 45 to 135 degrees). The fixed focal length network performs slightly better with matching images but loses precision with test images of different focal lengths. In contrast, the focal length-independent network demonstrates the capability to handle such differences. We argue that the significantly enhanced flexibility of the latter justifies the minor performance degradation in practical, real-world scenarios.

\nparagraph{Influence of Camera Rotation Angles.}
The range of pitch and roll angles the network is able to handle is determined by the variations of angles seen during training. We train the network for ZInD using a angle variation of $\pm 22^{\circ}$ and test with images where the range of variation increases from $0^{\circ}$ to $20^{\circ}$ in steps of $\pm 5^{\circ}$. \Cref{tab:angle} shows only a slight increase in median errors for rotation and translation as the rotation angles in test images increase. 
The chosen ranges represent a significant coverage and offer substantial improvements over existing methods.
In indoor scenes, larger camera tilt often results in images dominated by the ceiling or floor, rendering them unsuitable for semantic matching. 


\begin{table}[!t]
\begin{minipage}{0.53\linewidth}
\caption{The fixed FoV network excels at a specific angle but loses precision with changing FoV. The proposed network remains stable.}
\renewcommand{\arraystretch}{1.1}
\centering
\resizebox{.9\columnwidth}{!}{%
    \begin{tabular}{cccccc}
\toprule
 \begin{tabular}[c]{@{}c@{}}Input\\FoV\end{tabular} & \begin{tabular}[c]{@{}c@{}}Train\\FoV\end{tabular} & \begin{tabular}[c]{@{}c@{}}\textless{}1m med\\terr (cm)\end{tabular} & \begin{tabular}[c]{@{}c@{}}\textless{}1m med\\rerr (deg)\end{tabular} & \begin{tabular}[c]{@{}c@{}}10cm\\(\%)\end{tabular} & \begin{tabular}[c]{@{}c@{}}1m\\(\%)\end{tabular} \\
\midrule
\multirow{2}{*}{90°} & 90° & \textbf{14.31} & \textbf{2.05} & \textbf{26.88} & \textbf{91.90}  \\
& 45 - 135° & 16.11 & 2.20 & 21.81 & 91.45 \\
\midrule
\midrule
\multirow{2}{*}{80°} & 90° & 37.51 & 2.37 & 1.67 & 90.14  \\
&  45 - 135° & \textbf{16.75} & \textbf{2.19} & \textbf{19.30} & \textbf{90.50} \\
\bottomrule
\end{tabular}
}
\label{tab:spvloc-fov}
\end{minipage}%
\hfill
\begin{minipage}{0.44\linewidth}
\caption{Effect of increasing angle variation in test data.}
\renewcommand{\arraystretch}{1.1}
\centering
\resizebox{\columnwidth}{!}{%
    \begin{tabular}{cccccc}
 \toprule
 ~ & \multicolumn{5}{c}{Roll/Pitch Variation} \\
\cline{2-6}
Metric & $\pm 0^{\circ}$ & $\pm 5^{\circ}$ & $\pm 10^{\circ}$ & $\pm 15^{\circ}$ & $\pm 20^{\circ}$ \\
\midrule
<1m med. terr & 14.46 & 14.88 & 14.88 & 15.21 & 15.91  \\
<1m med. rerr & ~2.35 & ~2.36 & ~2.38 & ~2.51 & ~2.60  \\
10cm (\%) & 24.64 & 23.94 & 24.46 & 23.69 & 22.74 \\
1m (\%) & 91.45 & 91.47 & 91.99 & 92.38 & 92.60 \\
\bottomrule
\end{tabular}
}
\label{tab:angle}
\end{minipage}
\end{table}

\nparagraph{Limitations.}
In large spaces with repeating room layouts, the approach's effectiveness may be constrained by the limited level of detail in the semantic reference model. 
However, enhancing the S3D format by adding more semantic classes and structural information, such as staircases or sanitary installations, could reduce ambiguities.

\nparagraph{Timing.}
We split the measured time into scene sampling/preparation time, including position sampling, panorama rendering and encoding, and inference/query time, including query image encoding, feature correlation with bounding box classification, and the \textit{Pose} task head, but no refinement.
\Cref{table:timing} shows that our method exhibits a considerably lower query time than LASER \cite{Min2022} and PF-net \cite{Karkus2018}. This difference places our approach close to real-time for moderately sized scenes. Currently, scene sampling time is slower than \cite{Min2022}, mainly due to rendering on the CPU without parallelization \cite{Li2018}. Parallelizing this process using GPU or CPU could significantly accelerate scene preparation.
\begin{table}[!t]
    \caption{\textbf{Timing.} Performance on ZInD using a NVIDIA Tesla V100, following the \textit{base} configuration outlined in \cref{tab:spvloc-ablation}, and compared against results from \cite{Min2022}.}
    \renewcommand{\arraystretch}{1.1}
    \centering
    \resizebox{.7\columnwidth}{!}{%
        \begin{tabular}{ccccccc} 
\cline{1-3}\cline{5-6}
Method & Sampling Time (s) & Query Fps & ~ ~ ~ & Query Module & Time (ms) \\ 
\cline{1-3}\cline{5-6}
PF-net\cite{Karkus2018}  & 48.95  $\pm$ 38.95 & 5.06 $\pm$ 1.77  &  & Eff.Net V2 S & 4.4 \\
LASER\cite{Min2022}  & \bft{0.97} $\pm$ 1.09 & 8.31 $\pm$ 0.64 &  & Correlation & 35.7  \\
Ours  & 1.66 $\pm$  1.06 & \bft{28.63} $\pm$  11.54 &  & Pose & 1.1  \\
\cline{1-3}\cline{5-6}
\end{tabular}
    }
    \label{table:timing}
\end{table}

\section{Conclusion}
This paper introduces a method for scene-independent model-based 6D localization for indoor scenes, involving a novel method for multimodal image matching (panorama to perspective, RGB to semantic). The matching and retrieval is efficient and scalable under sparse reference sampling. The localization accuracy and inference speed outperform state of the art methods and the inclusion of the 3D-model, reduces ambiguities to precisely estimate 6D poses. Future work  involves combining localization and image analysis to enhance digital building models or exploring applications in augmented reality scenarios.

\nparagraph{Acknowledgements.}
This work is partially funded by the German Federal Ministry of Economic Affairs and Climate Action (BIMKIT, 01MK21001H) and the German  Federal Ministry for Digital and Transport (EConoM, 19OI22009C).
%
%
\bibliographystyle{splncs04}
\bibliography{main}

\begin{thebibliography}{10}
\providecommand{\url}[1]{\texttt{#1}}
\providecommand{\urlprefix}{URL }
\providecommand{\doi}[1]{https://doi.org/#1}

\bibitem{Acharya2019}
Acharya, D., Khoshelham, K., Winter, S.: {BIM-PoseNet: Indoor camera localisation using a 3D indoor model and deep learning from synthetic images}. ISPRS journal of photogrammetry and remote sensing  \textbf{150},  245--258 (2019)

\bibitem{Acharya2019bim}
Acharya, D., Ramezani, M., Khoshelham, K., Winter, S.: Bim-tracker: A model-based visual tracking approach for indoor localisation using a 3d building model. ISPRS Journal of Photogrammetry and Remote Sensing  \textbf{150},  157--171 (2019)

\bibitem{Acharya2023}
Acharya, D., Tatli, C.J., Khoshelham, K.: Synthetic-real image domain adaptation for indoor camera pose regression using a 3d model. ISPRS Journal of Photogrammetry and Remote Sensing  \textbf{202},  405--421 (2023)

\bibitem{Acharya2022}
Acharya, D., Tennakoon, R., Muthu, S., Khoshelham, K., Hoseinnezhad, R., Bab-Hadiashar, A.: {Single-image localisation using 3D models: Combining hierarchical edge maps and semantic segmentation for domain adaptation}. Automation in Construction  \textbf{136},  104--152 (2022)

\bibitem{Agarwal2011}
Agarwal, S., Furukawa, Y., Snavely, N., Simon, I., Curless, B., Seitz, S.M., Szeliski, R.: {Building rome in a day}. Communications of the ACM  \textbf{54}(10),  105--112 (2011)

\bibitem{Ammirato2018}
Ammirato, P., Fu, C.Y., Shvets, M., Kosecka, J., Berg, A.C.: {Target driven instance detection}. arXiv preprint arXiv:1803.04610  (2018)

\bibitem{Arandjelovic2016}
Arandjelovic, R., Gronat, P., Torii, A., Pajdla, T., Sivic, J.: Netvlad: Cnn architecture for weakly supervised place recognition. In: Proc.~IEEE/CVF Conference on Computer Vision and Pattern Recognition (CVPR) (2016)

\bibitem{Cambeiro2023}
Cambeiro, A.B., Trzeciakiewicz, M., Hilsmann, A., Eisert, P.: Automatic reconstruction of semantic 3d models from 2d floor plans. In: Proc. Int. Conf. on Machine Vision Applications (MVA). Hamamatsu, Japan (July 2023)

\bibitem{Chen2024}
Chen, C., Wang, R., Vogel, C., Pollefeys, M.: {F3Loc: Fusion and Filtering for Floorplan Localization}. In: Proc.~IEEE/CVF Conference on Computer Vision and Pattern Recognition (CVPR) (June 2024)

\bibitem{Cruz2021}
Cruz, S., Hutchcroft, W., Li, Y., Khosravan, N., Boyadzhiev, I., Kang, S.B.: {Zillow indoor dataset: Annotated floor plans with 360deg panoramas and 3d room layouts}. In: Proc.~IEEE/CVF Conference on Computer Vision and Pattern Recognition (CVPR). pp. 2133--2143 (2021)

\bibitem{Dellaert1999}
Dellaert, F., Fox, D., Burgard, W., Thrun, S.: Monte carlo localization for mobile robots. In: Proceedings IEEE International Conference on Robotics and Automation. vol.~2, pp. 1322--1328 (1999)

\bibitem{Ding2019}
Ding, M., Wang, Z., Sun, J., Shi, J., Luo, P.: {CamNet: Coarse-to-fine retrieval for camera re-localization}. In: Proc.~IEEE/CVF Conference on Computer Vision and Pattern Recognition (CVPR). pp. 2871--2880 (2019)

\bibitem{Fernandez2020}
Fernandez-Labrador, C., Facil, J.M., Perez-Yus, A., Demonceaux, C., Civera, J., Guerrero, J.J.: {Corners for layout: End-to-end layout recovery from 360 images}. IEEE Robotics and Automation Letters  \textbf{5}(2),  1255--1262 (2020)

\bibitem{He2012}
He, B., Chen, Z., Li, Y.: {Calibration method for a central catadioptric-perspective camera system}. JOSA A  \textbf{29}(11),  2514--2524 (2012)

\bibitem{He2016}
He, K., Zhang, X., Ren, S., Sun, J.: Deep residual learning for image recognition. In: Proc.~IEEE/CVF Conference on Computer Vision and Pattern Recognition (CVPR). pp. 770--778 (2016)

\bibitem{Howard2022}
Howard-Jenkins, H., Prisacariu, V.A.: {LaLaLoc++: Global Floor Plan Comprehension for Layout Localisation in Unvisited Environments}. In: Proc.~European Conference on Computer Vision (ECCV). pp. 693--709. Springer (2022)

\bibitem{Howard2021}
Howard-Jenkins, H., Ruiz-Sarmiento, J.R., Prisacariu, V.A.: {LaLaLoc: Latent Layout Localisation in Dynamic, Unvisited Environments}. In: Proc.~IEEE/CVF International Conference on Computer Vision (ICCV) (2021)

\bibitem{Janik2021}
Janik, M., Gard, N., Hilsmann, A., Eisert, P.: {Zero in on Shape: A Generic 2D-3D Instance Similarity Metric Learned from Synthetic Data}. In: Proc. IEEE International Conference on Image Processing (ICIP). pp. 2638--2642. IEEE (2021)

\bibitem{Karkus2018}
Karkus, P., Hsu, D., Lee, W.S.: Particle filter networks with application to visual localization. In: Conference on robot learning. pp. 169--178. PMLR (2018)

\bibitem{Kendall2017}
Kendall, A., Cipolla, R.: Geometric loss functions for camera pose regression with deep learning. In: Proc.~IEEE/CVF Conference on Computer Vision and Pattern Recognition (CVPR). pp. 5974--5983 (2017)

\bibitem{Kendall2018}
Kendall, A., Gal, Y., Cipolla, R.: Multi-task learning using uncertainty to weigh losses for scene geometry and semantics. In: Proc.~IEEE/CVF Conference on Computer Vision and Pattern Recognition (CVPR). pp. 7482--7491 (2018)

\bibitem{Kendall2015}
Kendall, A., Grimes, M., Cipolla, R.: Posenet: A convolutional network for real-time 6-dof camera relocalization. In: Proc.~IEEE/CVF International Conference on Computer Vision (ICCV). pp. 2938--2946 (2015)

\bibitem{Kim2024}
Kim, J., Jeong, J., Kim, Y.M.: Fully geometric panoramic localization. In: Proc.~IEEE/CVF Conference on Computer Vision and Pattern Recognition (CVPR). pp. 20827--20837 (2024)

\bibitem{Li2018}
Li, T.M., Aittala, M., Durand, F., Lehtinen, J.: {Differentiable Monte Carlo Ray Tracing through Edge Sampling}. ACM Trans. Graph. (Proc. SIGGRAPH Asia)  \textbf{37}(6),  222:1--222:11 (2018)

\bibitem{Lin2017}
Lin, T.Y., Goyal, P., Girshick, R., He, K., Doll{\'a}r, P.: {Focal loss for dense object detection}. In: Proc.~IEEE/CVF International Conference on Computer Vision (ICCV). pp. 2980--2988 (2017)

\bibitem{Liu2015}
Liu, C., Schwing, A., Kundu, K., Urtasun, R., Fidler, S.: Rent3d: Floor-plan priors for monocular layout estimation. In: Proc.~IEEE/CVF Conference on Computer Vision and Pattern Recognition (CVPR) (2015)

\bibitem{Liu2017}
Liu, L., Li, H., Dai, Y.: {Efficient global 2d-3d matching for camera localization in a large-scale 3d map}. In: Proc.~IEEE/CVF Conference on Computer Vision and Pattern Recognition (CVPR). pp. 2372--2381 (2017)

\bibitem{Lv2021}
Lv, X., Zhao, S., Yu, X., Zhao, B.: Residential floor plan recognition and reconstruction. In: Proc.~IEEE/CVF Conference on Computer Vision and Pattern Recognition (CVPR). pp. 16717--16726 (June 2021)

\bibitem{Mercier2021}
Mercier, J.P., Garon, M., Giguere, P., Lalonde, J.F.: {Deep Template-Based Object Instance Detection}. In: Proc. WACV. pp. 1507--1516 (January 2021)

\bibitem{Min2022}
Min, Z., Khosravan, N., Bessinger, Z., Narayana, M., Kang, S.B., Dunn, E., Boyadzhiev, I.: {Laser: Latent space rendering for 2d visual localization}. In: Proc.~IEEE/CVF Conference on Computer Vision and Pattern Recognition (CVPR). pp. 11122--11131 (2022)

\bibitem{Naseer2017}
Naseer, T., Burgard, W.: Deep regression for monocular camera-based 6-dof global localization in outdoor environments. In: Proc.~International Conference on Intelligent Robot Systems (IROS). pp. 1525--1530. IEEE (2017)

\bibitem{Orhan2021}
Orhan, S., Ba{\c{s}}tanlar, Y.: {Efficient search in a panoramic image database for long-term visual localization}. In: Proc.~IEEE/CVF Conference on Computer Vision and Pattern Recognition (CVPR). pp. 1727--1734 (2021)

\bibitem{Orhan2022}
Orhan, S., Guerrero, J.J., Ba\c{s}tanlar, Y.: {Semantic Pose Verification for Outdoor Visual Localization With Self-Supervised Contrastive Learning}. In: Proc.~IEEE/CVF Conference on Computer Vision and Pattern Recognition (CVPR) Workshops. pp. 3989--3998 (June 2022)

\bibitem{Park2021}
Park, S., Kim, H.: 3dplannet: generating 3d models from 2d floor plan images using ensemble methods. Electronics  \textbf{10}(22), ~2729 (2021)

\bibitem{Sarlin2021}
Sarlin, P.E., Unagar, A., Larsson, M., Germain, H., Toft, C., Larsson, V., Pollefeys, M., Lepetit, V., Hammarstrand, L., Kahl, F., et~al.: {Back to the feature: Learning robust camera localization from pixels to pose}. In: Proc.~IEEE/CVF Conference on Computer Vision and Pattern Recognition (CVPR). pp. 3247--3257 (2021)

\bibitem{Shi2023}
Shi, Z., Shi, H., Yang, K., Yin, Z., Lin, Y., Wang, K.: {PanoVPR: Towards Unified Perspective-to-Equirectangular Visual Place Recognition via Sliding Windows across the Panoramic View}. In: 2023 IEEE 26th International Conference on Intelligent Transportation Systems (ITSC). pp. 1333--1340 (2023)

\bibitem{Sturm2002}
Sturm, P.: {Mixing catadioptric and perspective cameras}. In: Proceedings of the IEEE Workshop on Omnidirectional Vision 2002. Held in conjunction with ECCV'02. pp. 37--44. IEEE (2002)

\bibitem{Taira2018}
Taira, H., Okutomi, M., Sattler, T., Cimpoi, M., Pollefeys, M., Sivic, J., Pajdla, T., Torii, A.: {InLoc: Indoor visual localization with dense matching and view synthesis}. In: Proc.~IEEE/CVF Conference on Computer Vision and Pattern Recognition (CVPR). pp. 7199--7209 (2018)

\bibitem{Tateno2018}
Tateno, K., Navab, N., Tombari, F.: {Distortion-aware convolutional filters for dense prediction in panoramic images}. In: Proc.~European Conference on Computer Vision (ECCV). pp. 707--722 (2018)

\bibitem{Torii2015}
Torii, A., Arandjelovic, R., Sivic, J., Okutomi, M., Pajdla, T.: {24/7 place recognition by view synthesis}. In: Proc.~IEEE/CVF Conference on Computer Vision and Pattern Recognition (CVPR). pp. 1808--1817 (2015)

\bibitem{Turkoglu2021}
T{\"{u}}rko\u{g}lu, M.{\"{O}}., Brachmann, E., Schindler, K., Brostow, G., Monszpart, A.: {Visual Camera Re-Localization Using Graph Neural Networks and Relative Pose Supervision}. In: Proc.~International Conference on 3D Vision (3DV) (2021)

\bibitem{Wang2015}
Wang, S., Fidler, S., Urtasun, R.: Lost shopping! monocular localization in large indoor spaces. In: Proc.~IEEE/CVF International Conference on Computer Vision (ICCV). pp. 2695--2703 (2015)

\bibitem{Xia2022}
Xia, J., Gong, J.: Precise indoor localization with 3d facility scan data. Computer-Aided Civil and Infrastructure Engineering  \textbf{37}(10),  1243--1259 (2022)

\bibitem{Zamir2010}
Zamir, A.R., Shah, M.: {Accurate image localization based on google maps street view}. In: Proc.~European Conference on Computer Vision (ECCV). pp. 255--268. Springer, Heraklion, Greece (Sep 2010)

\bibitem{Zhang2021}
Zhang, C., Budvytis, I., Liwicki, S., Cipolla, R.: {Lifted semantic graph embedding for omnidirectional place recognition}. In: Proc.~International Conference on 3D Vision (3DV). pp. 1401--1410. IEEE (2021)

\bibitem{Sheng2020}
Zheng, J., Zhang, J., Li, J., Tang, R., Gao, S., Zhou, Z.: {Structured3D: A large photo-realistic dataset for structured 3d modeling}. In: Proc.~European Conference on Computer Vision (ECCV). pp. 519--535. Springer, Glasgow, UK (Aug 2020)

\end{thebibliography}
\clearpage 
\thispagestyle{plain}


\maketitlesupplementary 
\setcounter{section}{0}

\renewcommand{\thesection}{\Alph{section}}
\setcounter{subsection}{0}
\renewcommand{\thesubsection}{\Alph{section}.\arabic{subsection}}
\setcounter{subsubsection}{0} 
\noindent This supplementary material provides additional insights about SPVLoc. \Cref{sec:supl_more_experiments} includes two additional experiments using the S3D dataset \cite{Sheng2020}.
\Cref{sec:supl_visual} presents more visual results on ZinD dataset \cite{Cruz2021} and examines specific aspects of SPVLoc. \Cref{sec:supl_details} discusses additional implementation details. 

\section{Additional Experiments}
\label{sec:supl_more_experiments}
\subsection{Refinement for a different Aspect Ratio} 
\label{subsec:aspect_ratio}
\begin{table}[b]
\caption{\textbf{Evaluation of SPVLoc on perspective images from S3D \cite{Sheng2020}.} We compare \textit{furnished} and \textit{empty} configuration. We list median errors for 2D localization (translation error terr\textsubscript{xy} in cm, rotation error rerr\textsubscript{yaw} in degrees) and 6D localization (terr\textsubscript{xyz} in cm, rerr denoting 3D rotation difference in degrees) for instances localized within 1m. We also report recall at various translation accuracy levels, recall for inliers ($<$1m and $<$30°), and top-3 recall at 1m.  }
\renewcommand{\arraystretch}{1.1}
\centering
\resizebox{1\columnwidth}{!}{%
    \begin{tabular}{ccccccccccc} 
\toprule
Query &  \begin{tabular}[c]{@{}c@{}}Aspect \vspace{-0.3em}\\ratio \end{tabular} & Furniture ~ & \begin{tabular}[c]{@{}c@{}}Localization \vspace{-0.3em}\\mode \end{tabular} & \begin{tabular}[c]{@{}c@{}}\textless{}1m med\vspace{-0.3em}\\terr (cm)\end{tabular} & \begin{tabular}[c]{@{}c@{}}\textless{}1m med\vspace{-0.3em}\\rerr (deg)\end{tabular} & \begin{tabular}[c]{@{}c@{}}10cm\vspace{-0.3em}\\(\%)\end{tabular} & \begin{tabular}[c]{@{}c@{}}50cm\vspace{-0.3em}\\(\%)\end{tabular} & \begin{tabular}[c]{@{}c@{}}1m\vspace{-0.3em}\\(\%)\end{tabular} & \begin{tabular}[c]{@{}c@{}}1m \& 30°\vspace{-0.3em}\\(\%)\end{tabular} & \begin{tabular}[c]{@{}c@{}}top-3\vspace{-0.3em}\\1m(\%)\end{tabular} \\ 
\hline
\multirow{4}{*}{\begin{tabular}[c]{@{}c@{}}Perspective\\80° FoV~~~\end{tabular}} &
\multirow{4}{*}{~~16:9~~} &
\multirow{2}{*}{empty} &
2D & 11.70 & 1.14 & 35.99 & 84.06 & 86.73 & 86.25 & 94.79 \\
~& ~& ~& 6D & 13.09 & 1.50 & 28.77 & 83.97 & 86.73 & 86.21 & 94.79 \\
\cline{3-11}
 &
\multirow{2}{*}{~} &
\multirow{2}{*}{furnished} &
2D &  12.86 & 1.26 & 30.55 & 81.58 & 84.50 & 84.20 & 93.36 \\
~& ~& ~& 6D & 14.30 & 1.62 & 23.54 & 81.48 & 84.50 & 84.18 & 93.36 \\
\bottomrule
\end{tabular}
}
\label{tab:s3d_supplementary}
\vspace{-1em}
\end{table}
Besides panoramic renderings, S3D \cite{Sheng2020} also comprises high-resolution perspective images in a 16:9 format with 80 degree horizontal field of view. While we demonstrated a pipeline with perspective training images re-sampled from the 21835 panorama images in the main paper, this experiment showcases the usage of the perspective images coming with S3D during training and testing.

\Cref{tab:s3d_supplementary} lists the results for the test sets, including 6405 photo-realistically rendered perspective images from S3D test scenes, both with and without furniture. 
We use our pretrained network, as in \cref{tab:sota} of the main paper, and refine it for 10 more epochs with the S3D perspective images. During training, both \textit{empty} and \textit{furnished} images were used randomly. The images are zero-padded above and below to fit the square input of the network. 
With over 85\% (6D, average of empty and furnished) of the images correctly localized within a 1-meter threshold as the top-1 match and over 94\% as the top-3 match, the network demonstrates its capability to handle the 16:9 test images efficiently. More than 26\% of the images are localized correctly within a 10 cm threshold. 

To our knowledge, the only other work that provides localization results for the perspective images of S3D is F\textsuperscript{3}Loc \cite{Chen2024}. This method performs 2D localization and relies solely on a 2D floor plan occupancy without semantic information. \Cref{tab:suppl_f3loc} compares our method with the single-frame variant of F\textsuperscript{3}Loc. Although omitting semantic and height information can be advantageous in certain situations, it significantly reduces the recall rate. The table demonstrates how beneficial using this information is for localization, with nearly four times higher 1m \& 30° recall and even greater improvements for 50cm and 10cm recall.

\begin{table}[t]
\caption{\textbf{Comparison to the single-frame variant of F\textsuperscript{3}Loc}\cite{Chen2024} using \textit{furnished} perspective images from S3D. We report recall at different accuracy levels for 2D localization. Results for F\textsuperscript{3}Loc are taken from \cite{Chen2024}. }
\renewcommand{\arraystretch}{1.1}
\centering
\resizebox{.5\columnwidth}{!}{%
    \begin{tabular}{ccccccc}
\toprule
 Method & \begin{tabular}[c]{@{}c@{}}Uses\\semantics\end{tabular} & \begin{tabular}[c]{@{}c@{}}Reference\\model\end{tabular} & \begin{tabular}[c]{@{}c@{}}10cm\\(\%)\end{tabular} & \begin{tabular}[c]{@{}c@{}}50cm\\(\%)\end{tabular} & \begin{tabular}[c]{@{}c@{}}1m\\(\%)\end{tabular} & \begin{tabular}[c]{@{}c@{}}1m \& 30°\vspace{-0.3em}\\(\%)\end{tabular} \\
\midrule
F\textsuperscript{3}Loc\cite{Chen2024}  & no & 2D & 1.50  & 14.60  & 22.40 & 21.30  \\
\midrule
Ours & yes & 3D & \textbf{30.55} & \textbf{81.58} & \textbf{84.50} & \textbf{84.20} \\
\bottomrule
\end{tabular}
}
\label{tab:suppl_f3loc}
\vspace{-1em}
\end{table}

\begin{figure}[b!]
  \centering 
\includegraphics[width=0.24\linewidth]{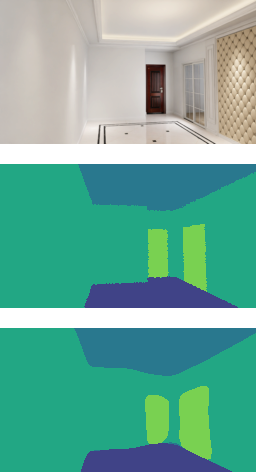}
\includegraphics[width=0.24\linewidth]{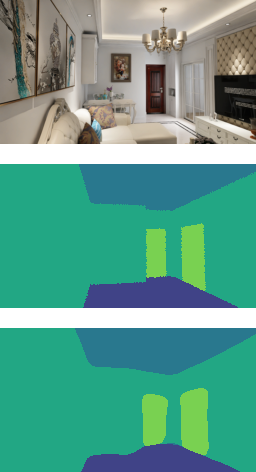}%
\hfill
\includegraphics[width=0.24\linewidth]{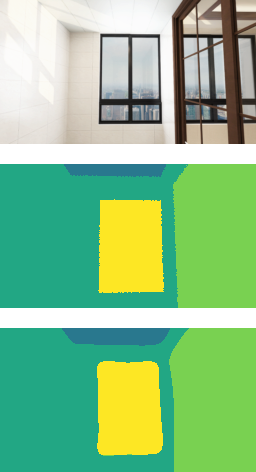}
\includegraphics[width=0.24\linewidth]{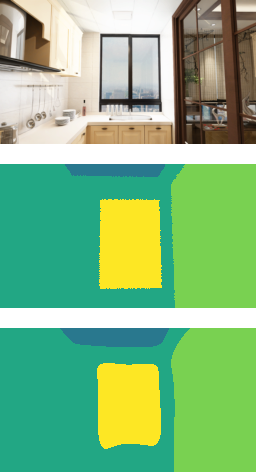}%
  \caption{\textbf{S3D examples with accurate localization for both \textit{empty} and \textit{furnished} images (terr < 25cm, rerr < 2°).} Input image in the first row, rendering with top-1 pose in the second, and view decoder output in the third.}
  \label{fig:furniture_success}
\end{figure}
\subsection{Robustness against Furniture} 
\begin{figure}
  \centering
    \begin{subfigure}{0.48\linewidth}
\includegraphics[width=0.49\linewidth]{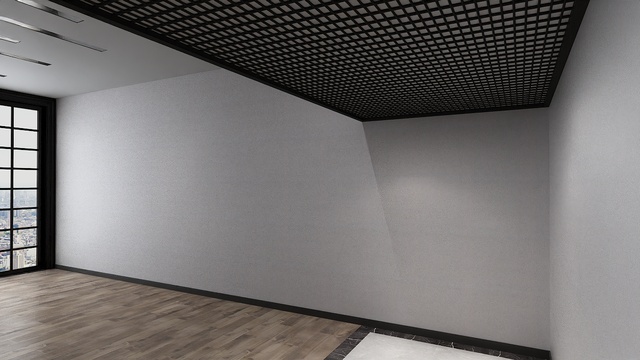}%
\hfill
\includegraphics[width=0.49\linewidth]{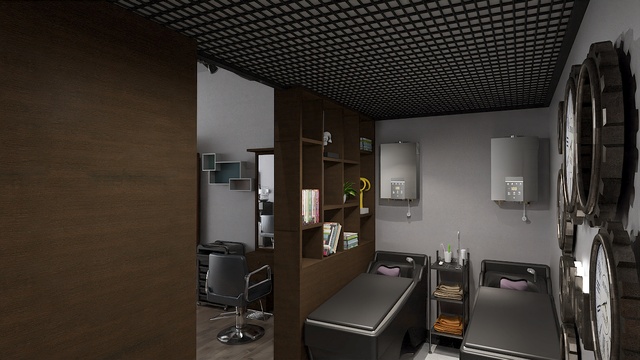}%
\caption{additional wall modelled as furniture}
\end{subfigure}
\hfill
    \begin{subfigure}{0.48\linewidth}
\includegraphics[width=0.49\linewidth]{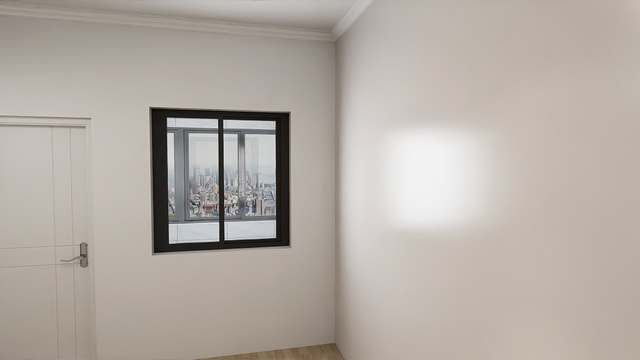}%
\hfill
\includegraphics[width=0.49\linewidth]{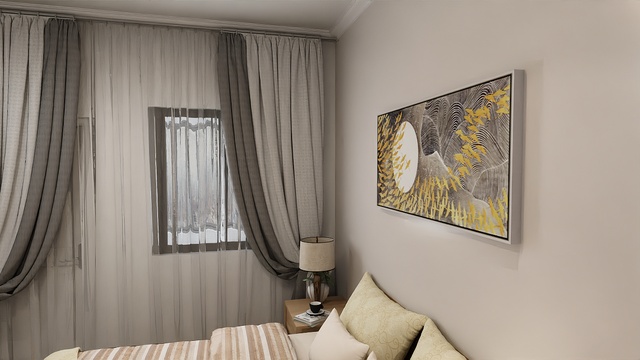}%
\caption{door behind curtain}
\end{subfigure}
  \caption{\textbf{Examples from S3D where the \textit{empty} image is localized accurately (terr < 25cm, rerr < 2°), but localization for the \textit{furnished} image fails.} In both cases important room structure is hidden behind furniture.}
  \label{fig:furniture_fail}
\end{figure}

S3D contains renderings of rooms in \textit{empty} and \textit{furnished} configuration. \Cref{tab:s3d_supplementary} compares the estimation results for \textit{empty} and \textit{furnished} test images, but otherwise equal camera poses. The results indicate, that the trained network is able to efficiently learn to ignore furniture in the images. This can also be seen by the output of the view decoder (the upper branch of \cref{fig:view_correlation} in the main paper) in \cref{fig:furniture_success}. 
The decoding of the input image is similar for the \textit{empty} and \textit{furnished} image and in both cases similar to the image rendered with the estimated pose. The network suppresses furniture very efficiently.
The slight decrease of around 2\% in 1m recall can mainly be attributed to furniture that obscures the visible room structure and makes matching difficult (\cref{fig:furniture_fail}). The decrease of around 5\% in 10cm recall can be explained, for instance, by wardrobes covering room edges, requiring the network to guess the underlying room structure.

\section{Visual Results}

\label{sec:supl_visual}

\begin{figure}[b!]
  \centering 
  \includegraphics[width=\linewidth]{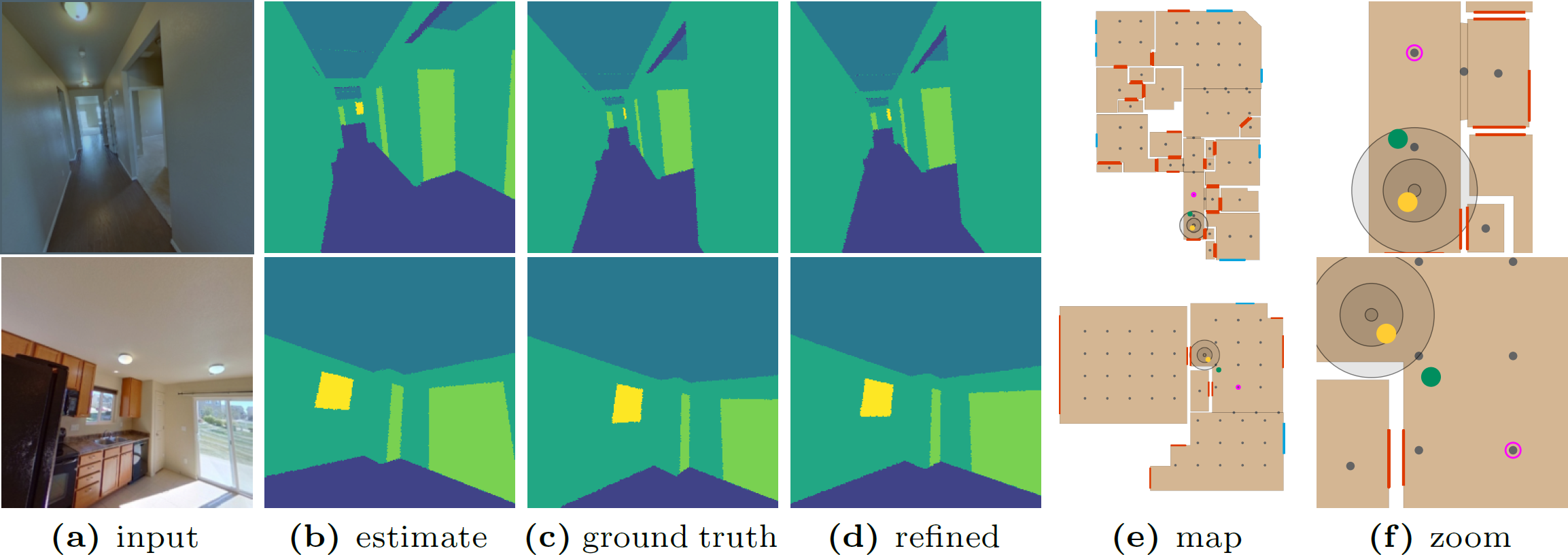}
  \caption{\textbf{Demonstration of pose improvement via an iterative refinement step.} On the map, the ground truth position is marked by gray circles with radii of 10cm, 50cm, and 100cm. The estimated position is represented by a green dot, the refined position by a yellow dot, and the position of the selected reference panorama by a pink dot. The grid size is 150cm and a local grid (Adv.) is used.}
  \label{fig:refinement}
\end{figure}

\subsection{Discussion of Refinement} 
The ablation study in the main paper showed how an optional refinement step leads to an improvement in localization recall and pose accuracy. In \cref{fig:refinement}, we show two examples, where the proposed refinement improves the estimated pose. 
The zoom in the map shows that the estimated pose is closer to the ground truth than the initial estimate. The relative pose offset between selected reference panorama position (pink) and the estimated position (green) is substantially larger than the distance between the estimated position (green) and the refined position (yellow).
As shown in \cref{tab:grid_size_quantitative} of the main paper, refinement is particularly useful when the distance between reference panorama positions is large, compensating for the cost of rendering an additional panorama during refinement. 

During refinement, the query is matched against a single panorama only, leading to fast inference.
Beyond the scope of this paper, our method could also be used for local tracking with a moving camera, where the panorama position is always located at the estimated position of the previous frame.

\begin{figure}[b!]
  \centering 
\includegraphics[width=0.95\linewidth]{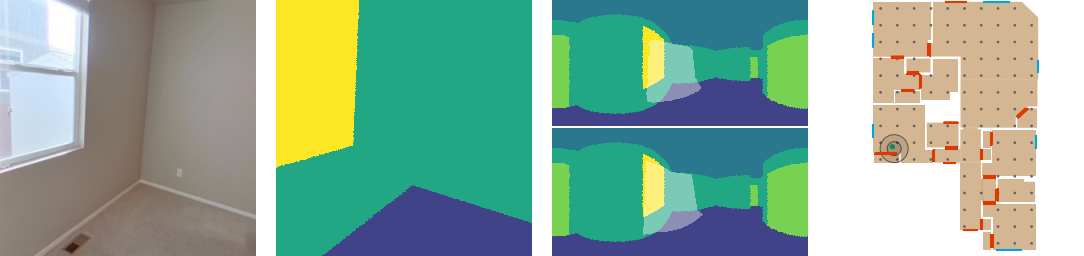} \\
\vspace{0.5em}
\includegraphics[width=0.95\linewidth]{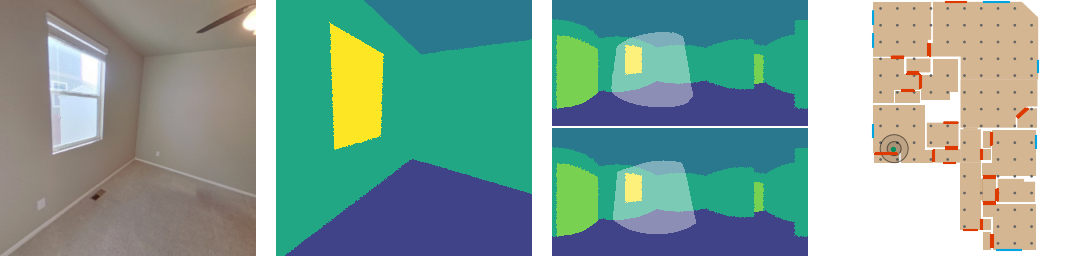} \\
\vspace{0.5em}
\includegraphics[width=0.95\linewidth]{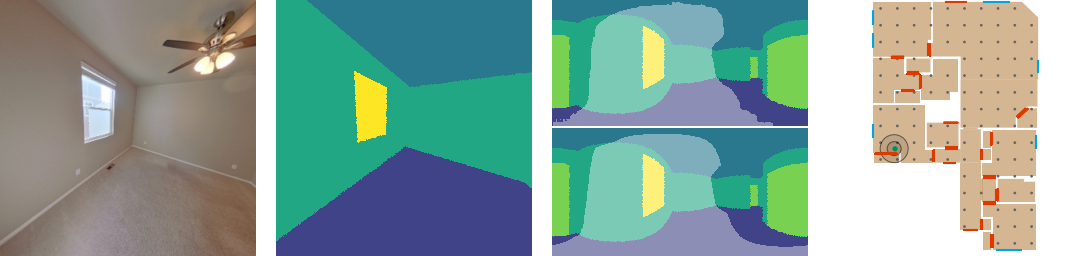} \\
  \caption{\textbf{Localization results for images with varying fields of view (60°, 90°, 120° from top to bottom) captured from the same position}. Input image in the first column, rendering with top-1 pose in the second column, comparison of estimated (top, white transparent overlay) and ground truth viewport mask (bottom) in the third column, the map in the fourth column. Ground truth position marked by gray circles (50cm and 100cm radius), estimated position by a green dot.}
  \label{fig:field_of_view_comparison}
\end{figure}

\subsection{Field of View Variation} 
\Cref{fig:field_of_view_comparison} shows an example of SPVLoc correctly estimating the pose of input images with different opening angles, as quantitatively presented in \cref{tab:spvloc-fov} in the main paper. Examining the estimated positions on the map reveals that all three images are correctly placed near the ground truth position, which remains consistent across all three instances. Additionally, it is evident that the view rendered from the top-1 camera pose closely matches the input image.
The third column presents a comparison of the viewport estimated with the \textit{Segment} head of the network (top, white transparent overlay) and the ground truth viewport (bottom), generated via perspective projection. By comparing viewport estimation, viewport ground truth, and the input image, we can observe that the network efficiently learns the perspective projection and establishes visual correspondences between the input and panorama. 
Importantly, to handle images with varying opening angles, the viewing angle of the camera must be inputted to the \textit{Pose} head. This implies that knowledge of basic sensor and lens properties is necessary.


\subsection{Discussion of Viewport Estimation} 
To evaluate how well a reference panorama corresponds to an input image, SPVLoc determines a classification score per reference panorama. The method employs anchor-based bounding box estimation \cite{Lin2017}, where each anchor is classified as either corresponding or not corresponding to the input image. The classification score for a reference panorama is the maximum score among all anchors.

\begin{figure}[p!]
  \centering 
\begin{subfigure}{\linewidth}
\centering 
\includegraphics[width=.32\linewidth]{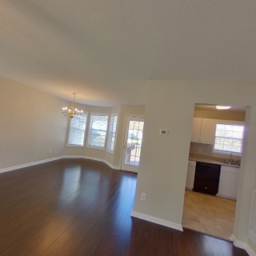}
\includegraphics[width=.32\linewidth]{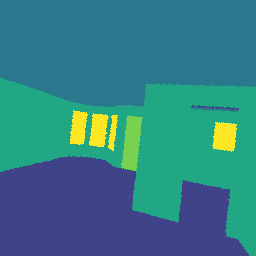} 
\includegraphics[width=.32\linewidth]{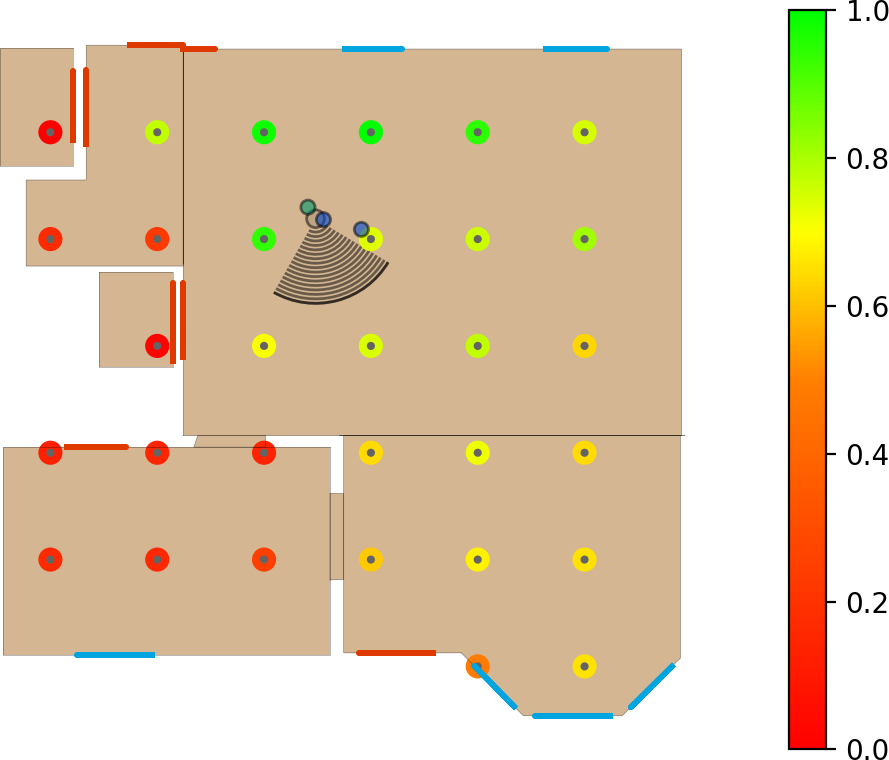}\\

\includegraphics[width=.32\linewidth]{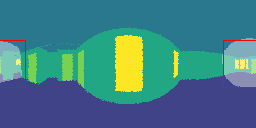} 
\includegraphics[width=.32\linewidth]{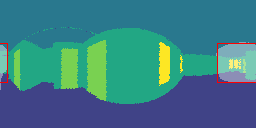}
\includegraphics[width=.32\linewidth]{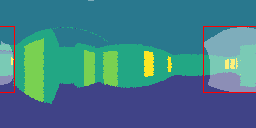}
\caption{Example 1}
\vspace{1.0em}
\end{subfigure}
\begin{subfigure}{\linewidth}
\centering 
\includegraphics[width=.32\linewidth]{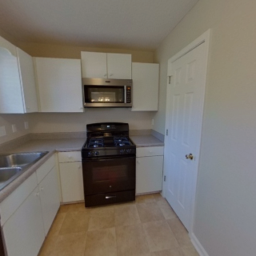} 
\includegraphics[width=.32\linewidth]{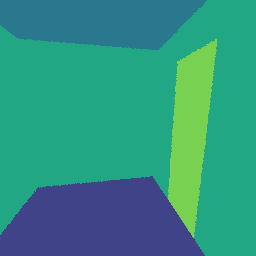} 
\includegraphics[width=.32\linewidth]{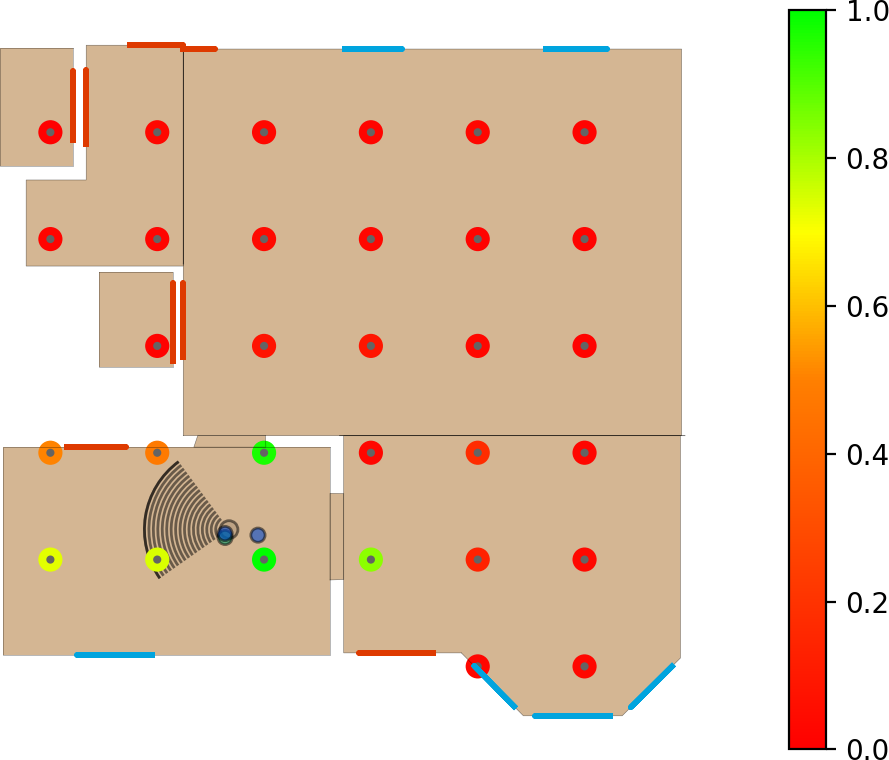}\\
\includegraphics[width=.32\linewidth]{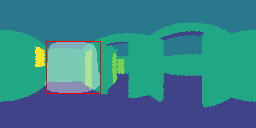} 
\includegraphics[width=.32\linewidth]{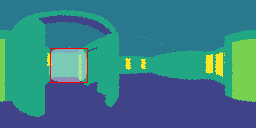}
\includegraphics[width=.32\linewidth]{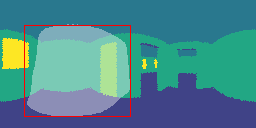}
\caption{Example 2}
\end{subfigure}
  \caption{\textbf{Viewport score overview.} For each example, the top row displays the input image on the left, the model rendered from the estimated top-1 pose in the middle, and a map where each reference position is color-coded relative to the estimated matching score. Here, the ground truth pose is shown with a stripe pattern.
  For each example, the second row shows the three panoramas with the highest matching scores, along with the estimated viewport displayed as a semi-transparent white overlay and a red bounding box.}
  \label{fig:map_overview}
\end{figure}

In \cref{fig:map_overview}, reference positions are color-coded based on their normalized classification score, ranging between 0 and 1, and displayed in a gradient from red to green. The ground truth camera direction is shown with an arc pattern originating from the ground truth position. The estimated top-3 locations are displayed by smaller transparent circles. For both examples, we show three reference panoramas with the highest classification scores. 
The exact pixel-based viewport estimated by the \textit{Segment} head is shown as a semi-transparent white overlay. We see that the reference panoramas achieving high scores capture the content of the input image to a large degree. For example, in Example 1, high scores are achieved for positions where the bay window as well as the wall opening in the direction of the kitchen can be seen in the panorama.

The selected reference panoramas show that the viewport of the camera image can be estimated from panorama positions with large in-between distances. Looking at the viewport in Example 2 (bottom row), the door as well as the adjoining wall are correctly detected by the viewport mask from different reference positions despite perspective distortion.

In conclusion, the depthwise-correlation used for the matching demonstrates its ability to incorporate finely-granulated details from the query encoder into the main network branch. It effectively distinguishes subtle distinctions in target images, maps from the photo to the semantic domain, and, in conjunction with the panoramas, successfully induces the correct 6D pose offset. While being based on deep template matching \cite{Mercier2021}, the ability of the operation to bridge the domain gap between semantic and real images as well as dealing with severe perspective distortion is one of the main novelties of our work.

\section{Further Details} 
\label{sec:supl_details}
\nparagraph{Design of the Matching Module.}
Our matching employs feature correlation, where several operations fuse backbone features ($F$ and $R$) from the two branches. Previous studies \cite{Ammirato2018, Mercier2021} support the necessity of our module design. The ablation study in \cite{Ammirato2018} shows improved matching by combining element-wise multiplication (there, referred to as cross-correlation) and subtraction with the $N \times 1 \times 1$ tensor (justifying $R^2$ and $R^3$). Building on this, the ablation study in \cite{Mercier2021} further proves that correlation with the $N \times 3 \times 3$ tensor and the concatenation of three tensors enhances matching accuracy (justifying $R^1$).

The increase of resolution of $R^*$ is attributed to bilinear scaling the features with a small factor. It facilitates linking output features with the pose head by maintaining spatial resolution before flattening. 

\nparagraph{Negative Samples in Training.} Negative samples are selected across the map, outside of radius $r_1$ around target. If the pose is sampled in a different room than the target image, both mask and target scores for all anchors are zero; in the same room, mask and bounding box are calculated as for positive samples, yielding larger offsets. The \textit{Pose} head is not evaluated for negative samples.

\nparagraph{FoV Scaling.} The input image is square, with the assumption that the horizontal and vertical FoV are the same. If an image with a different aspect ratio is used, we recommend zero-padding the centered image to square shape to achieve compatibility with the network (see \cref{subsec:aspect_ratio}). The input for the FoV scaling will then be the larger angle of both. To use the model with zero-padded images, this must be taken into account during training, e.g. by randomly adding symmetrical black bars of different widths at the edges of the image.

\nparagraph{Invariance to 3D-Model Inaccuracies.} While ZiND generally annotates ceiling height, it is not always modelled correctly due to inaccuracies and limitations of the data model, e.g., sloping roofs are not modelled in the S3D format, nor are they annotated in ZiND. During training, the network sees images, where mismatches occur and learns to deal with those imperfections e.g.\ by focusing on other visible elements like windows and doors. If height information is not available at inference time, we recommend using approximated fixed standard heights.

\nparagraph{Test Set Design and Comparison with LASER.}
We report slightly different results than those of LASER \cite{Min2022} as we conducted experiments using a test set designed to better evaluate pose estimation by requiring contextual information in the images. 
In contrast to the results of the original LASER experiment, our test images yield more inliers due to their richer semantic variation, albeit with slightly higher median translation errors due to computations over more inliers.
LASER's original experiment involved randomly sampling test yaw angles, resulting in images where the camera faced empty walls, making pose estimation impossible. The authors did not publish the specific angles used for each panorama, whereas we disclose all sampled angles to ensure the reproducibility of our dataset.

\end{document}